# Learning and Reasoning with Action-Related Places for Robust Mobile Manipulation

**Freek Stulp**                                                        STULP@CLMC.USC.EDU
*Computational Learning and Motor Control Lab*
*University of Southern California*
*3710 S. McClintock Avenue, Los Angeles, CA 90089, USA*

**Andreas Fedrizzi**                                                   FEDRIZZA@CS.TUM.EDU
**Lorenz Mösenlechner**                                                MOESENLE@CS.TUM.EDU
**Michael Beetz**                                                      BEETZ@CS.TUM.EDU
*Intelligent Autonomous Systems Group*
*Technische Universität München*
*Boltzmannstraße 3, D-85747 Garching bei München, Germany*

## Abstract

We propose the concept of Action-Related Place (ARPLACE) as a powerful and flexible representation of task-related place in the context of mobile manipulation. ARPLACE represents robot base locations not as a single position, but rather as a collection of positions, each with an associated probability that the manipulation action will succeed when located there. ARPLACEs are generated using a predictive model that is acquired through experience-based learning, and take into account the uncertainty the robot has about its own location and the location of the object to be manipulated.

When executing the task, rather than choosing one specific goal position based only on the initial knowledge about the task context, the robot instantiates an ARPLACE, and bases its decisions on this ARPLACE, which is updated as new information about the task becomes available. To show the advantages of this least-commitment approach, we present a transformational planner that reasons about ARPLACEs in order to optimize symbolic plans. Our empirical evaluation demonstrates that using ARPLACEs leads to more robust and efficient mobile manipulation in the face of state estimation uncertainty on our simulated robot.

## 1. Introduction

Recent advances in the design of robot hardware and software are enabling robots to solve increasingly complex everyday tasks. When performing such tasks, a robot must continually decide on its course of action, where a decision is *"a commitment to a plan or an action parameterization based on evidence and the expected costs and benefits associated with the outcome."* (Resulaj, Kiani, Wolpert, & Shadlen, 2009). This definition highlights the complexity of decision making. It involves choosing the *appropriate action* and *action parameterization*, such that *costs* are minimized and *benefits* are maximized. The robot must therefore be able to predict which costs and benefits will arise when executing an action. Furthermore, due to stochasticity and hidden state, the exact *outcome* of an action is not known in advance. The robot must therefore reason about *expected* outcomes, and be able to predict the probability of different outcomes for a given action and action parame-terization. Finally, a robot *commits* to decisions based on the current observable *evidence*, represented in its belief state. So if the evidence changes, the rationale for committing to a decision may no longer be valid. The robot therefore needs methods to efficiently reconsider decisions as the belief state changes during action execution, and possibly commit to another plan if necessary.

Mobile manipulation is a good case in point. Even the most basic mobile manipulation tasks, such as picking up an object from a table, require complex decision making. To pick up an object the robot must decide where to stand in order to pick up the object, which hand(s) to use, how to reach for it, which grasp type to apply, where to grasp, how much grasp force to apply, how to lift the object, how much force to apply to lift it, where to hold the object, and how to hold it. Such decision problems are complex as they depend on the specific task context, which consists of many task-relevant parameters. Furthermore, these decisions must be continually updated and verified, as the task context, or the robot's knowledge about the context, often changes during task execution.

Consequently, tasks of such complexity require not only robust hardware and low-level controllers, but also a least-commitment approach to making decisions, abstract planning capabilities, probabilistic representations, and principled ways of updating beliefs during task execution. In this article, we demonstrate how implementing these core AI topics contributes to the robustness and flexibility of our mobile manipulation platform.

The task-relevant decision we consider in this article is to which base position the robot should navigate to in order to perform a manipulation action. This decision alone presents several **challenges**, such as 1) successfully executing the reaching and manipulation action critically depends on the position of the base; 2) due to imperfect state-estimation, there is uncertainty in the position of the robot and the target object. As these positions are not known exactly, but are fundamental to successfully grasping the object, it is not possible to determine a single-best base position for manipulation; 3) the complete knowledge required to determine an appropriate base position is often not available initially, but rather acquired on-line during task execution.

Our **solution idea** to address these challenges is the concept of Action-Related Places (ARPLACE), a powerful and flexible representation of task-related place in the context of mobile manipulation. ARPLACE is represented as a probability mapping, which specifies the expected probability that the target object will be successfully grasped, given the positions of the target object and the robot

$$\text{ARPLACE}: \ \{ \ P(Success | \mathbf{f}_k^{rob}, \langle \hat{\mathbf{f}}^{obj}, \mathbf{\Sigma}^{obj} \rangle) \ \}_{k=1}^{K} \tag{1}$$

Here, the estimated position of the target object is represented as a multi-variate Gaussian distribution with mean $\hat{\mathbf{f}}^{obj}$ and covariance matrix $\mathbf{\Sigma}^{obj}$ [1]. The discrete set of robot positions $\{\mathbf{f}_k^{rob}\}_{k=1}^{K}$ should be thought of as possible base positions the robot considers for grasping, i.e. potential positions to navigate to. Typically, this set of positions is arranged in a grid, as in the exemplary ARPLACE depicted in Figure 1.

---

1. The feature vectors $\mathbf{f}^{rob}$ and $\mathbf{f}^{obj}$ contain the poses of the robot and the object relative to the table's edge. Details will be given in Section 3.1. Positions without uncertainty are denoted $\mathbf{f}$, and estimated positions with uncertainty as $\langle \hat{\mathbf{f}}, \mathbf{\Sigma} \rangle$.

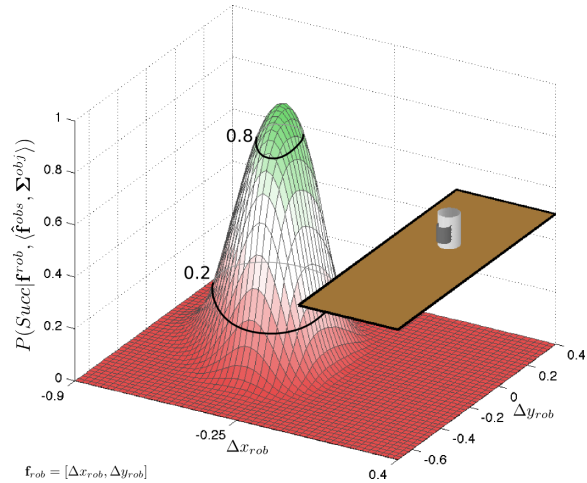

Figure 1: ARPlace: the probability of successful manipulation, given the current estimated object position $\langle \hat{\mathbf{f}}^{obj}_{cur}, \mathbf{\Sigma}^{obj}_{cur} \rangle$. The set of potential robot positions $\{\mathbf{f}^{rob}_k\}^K_{k=1}$ are arranged in a grid along the $x$ and $y$-axis, whereby each position leads to a different probability of successful grasping $P(Success|\mathbf{f}^{rob}_k, \langle \hat{\mathbf{f}}^{obj}_{cur}, \mathbf{\Sigma}^{obj}_{cur} \rangle)$. The black isolines represent grasp success probability levels of 0.2 and 0.8.

ARPlace has three important properties: 1) it models base places not as a single position $\mathbf{f}^{rob}$, but rather as a set of positions $\{\mathbf{f}^{rob}_k\}^K_{k=1}$, each with a different expectation of the success of the manipulation action; 2) it depends upon the estimated target object position, so updating $\hat{\mathbf{f}}^{obj}$ or $\mathbf{\Sigma}^{obj}$ during task execution thus leads to different probabilities in ARPlace; 3) using a probabilistic representation that takes into account uncertainty in the target object position leads to more robust grasping.

## 1.1 Example Scenario

In Figure 2, we present an **example scenario** that demonstrates how these properties of ARPlace address the challenges stated above, and supports decision-making during mobile manipulation. The images in the top row show the current situation of the robot from an outside view, while the images in the lower visualize the robot's internal ARPlace representation. The ARPlace is visualized with the colors red, white and green, which represent low, medium and high grasp success probabilities respectively. Grasp success probability levels of 0.2 and 0.8 are depicted as isolines, as in Figure 1.

In this scenario the robot's task is to clean the table. In Scene 1 the robot enters the kitchen and its vision system detects a cup. Because the robot is far away from the cup, the uncertainty arising from the vision-based pose estimation of the cup is high, indicated by the large circle around the cup in the lower left image. As the exact position of the cup is not known, it is not possible to determine a single-best base position for grasping the cup. The ARPlace representation takes this uncertainty into account by modeling the base position as a probability mapping.

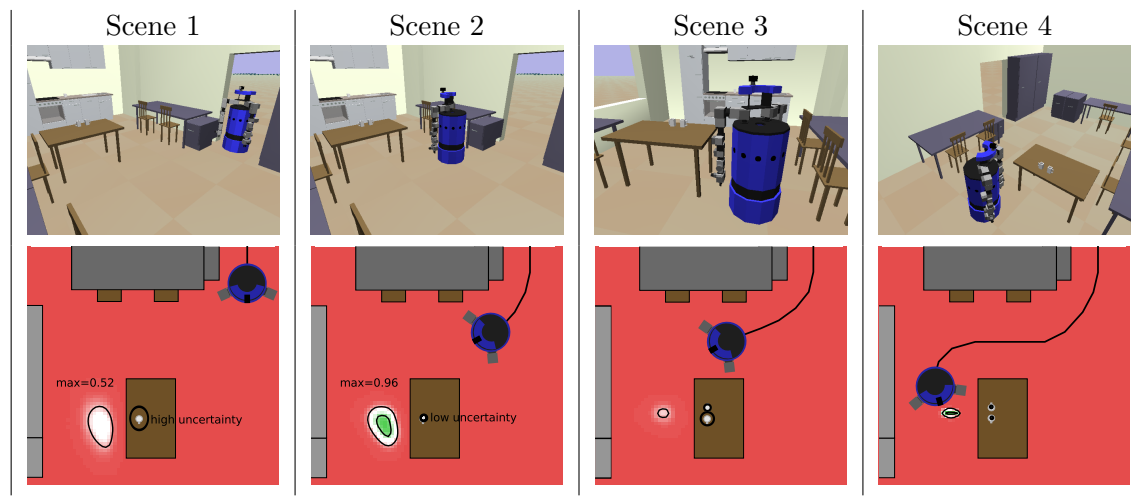

Figure 2: Example scenario.

In Scene 1 the ARPLACE distribution has low probabilities overall, with a maximum probability of grasp success of only 0.52. Note that although the initial uncertainty in the cup's position precludes the robot from determining a specific base position from which to reliably grasp the cup, the robot does know the general area to which it should navigate. During navigation, the robot is able to determine the position of the cup more accurately, as depicted in Scene 2. As new sensor data comes in, the robot refines the ARPLACE and therefore the ARPLACE in Scene 2 has much higher probabilities overall, with a maximum of 0.96.

In Scene 3 the robot has detected a second cup. Because grasping both cups at once from a single position is much more efficient than approaching two locations, the robot merges the two ARPLACEs for each cup into one ARPLACE representing the probability of successfully grasping *both* cups from a single position[2]. In Scene 4 further measurements helped to reduce pose estimation uncertainties of both cups. The maximum grasp success probability in the ARPLACE now reaches 0.97; sufficient for the robot to commit itself to a goal position and attempt to grasp both cups at once.

This scenario illustrates that real-world tasks can often not be planned from start to finish, as the initial knowledge is often not complete or accurate enough to determine an optimal goal position. So rather than committing to a particular base position early based only on the robot's initial knowledge about the task context, a robot instantiates an ARPLACE for a particular task context, and bases its decisions on this ARPLACE. Having the place concept instantiation **represented explicitly** during the course of action enables the robot to **reconsider and reevaluate these decisions on-line** whenever new information about the task context comes in. For instance, the decision to grasp both cups from one position in Scene 3 would not have been possible if the robot would have committed itself to a plan given its initial knowledge when only one cup was detected. Even if the environment is completely observable, dynamic properties can make a pre-planned optimal position suboptimal or unaccessible. A least-commitment implementation, where decisions are delayed until they must be taken is more flexible, and leads to more robust and efficient mobile manipulation. This will be demonstrated in the empirical evaluation.

---

2. Section 4.4.1 explains how ARPLACEs are merged to compute an ARPLACE for joint tasks.

## 1.2 Contributions, System Overview and Outline

The system overview for learning, computing, and reasoning with ARPLACEs is depicted in Figure 3. It also serves as an outline for the rest of this article.

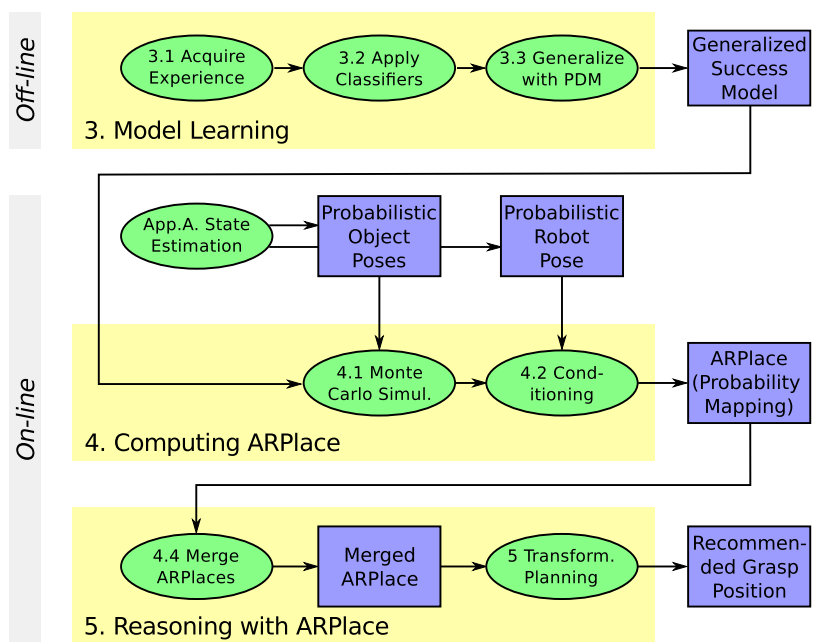

Figure 3: System Overview. Numbers refer to sections in this article. Green ovals represent algorithms and procedures, and blue rectangles the models that result from them. Procedures and models are briefly described in the contributions in this section; more detail is given throughout the article. Yellow rectangles cluster conceptually related procedures, and also delineate the different sections of this article.

The main contributions of this article are:

**Representing ARPlace – Section 1.** We propose ARPLACE as a flexible representation of place for least-commitment decision making in mobile manipulation.

**Model Learning – Section 3.** To generate an ARPLACE, the robot must be able to predict the outcome of an action for a given action parameterization. We propose a generic, off-line learning approach to acquiring a compact prediction model in two steps: 1) learn to predict whether an action will succeed for a given task parameterization. This is a supervised classification problem which we implement with Support Vector Machines (Sonnenburg, Raetsch, Schaefer, & Schoelkopf, 2006); 2) generalize over several task parameterizations by generalizing over the learned SVM classifiers, which we implement with Point Distribution Models (Cootes, Taylor, Cooper, & Graham, 1995). The resulting *success prediction model* enables the robot to predict whether a manipulation action for a given object position will succeed from a given base position [3].

**Generating ARPlace– Section 4.** We demonstrate how ARPLACEs are generated online, and take object position uncertainty into account through a Monte-Carlo simulation. Furthermore, the ARPLACE is conditioned on robot position uncertainty, which is thus also taken into account.

**Reasoning with ARPlace– Section 5.** We show how ARPLACE is integrated in a symbolic transformational planner, to automate decision-making with ARPLACEs. In particular, we consider a scenario that shows how ARPLACEs can be merged for joint manipulation tasks.

**Empirical Evaluation – Section 6.** We demonstrate how reasoning with ARPLACEs leads to more robust and efficient behavior on a simulated mobile manipulation platform.

Before turning to these contributions, we first compare our approach to related work in Section 2.

## 2. Related Work

Most state-of-the-art mobile manipulation platforms use sampling-based motion planners to solve manipulation problems (LaValle, 2006). Some of the advantages of using symbolic planning in general, and with ARPLACEs in particular, are: *1. Abstraction.* Representing and planning with abstract symbolic actions reduces the complexity of the planning problem. Although computational power is ever increasing, it is still intractable to solve extended tasks, such as preparing a meal (Beetz et al., 2008), with state-space search alone. *2. Least-commitment.* Friedman and Weld (1996) show that setting open conditions to abstract actions and later refining this choice to a particular concrete action can lead to exponential savings. Note that this principle has also be used to reduce the number of collision checks for building Probabilistic Roadmaps (Bohlin & Kavraki, 2000). *3. Modular replanning.* In symbolic planning, causal links between the actions are explicitly represented. That the robot navigates to the table *in order* to perform a grasping motion is not represented in a plan generated by a sampling-based motion planner. Therefore, during execution, motion planners cannot reconsider the appropriate base position as a decision in its own right, but must rather inefficiently replan the entire trajectory if the belief state changes. *4. Reflection.* The explicit symbolic representation of causality also allows a robot to reason about and reflect its own plans and monitor their execution, for instance to report reasons for plan failure: "I could not find the cup." or "I could not determine the position of the cup with sufficient accuracy to robustly perform the grasp." or "An obstacle was blocking my path.". It is not obvious how to achieve such introspection with motion planning methods.

---

3. By using experience-based learning, our approach can be applied to a variety of robots and environments. Once a model has been learned however, it is obviously specific to the environment in which the experience was generated. For instance, if only one table height is considered during data collection, the learned prediction model will be specific to that table height. If different table heights are used during experience collection, and the table height is included as a task-relevant parameter, the model should be able to generalize over table heights as well. We refer to Section 3.5 for a full discussion.

Also, in contrast to sampling based motion-planning, an ARPLACE itself does *not* generate trajectories or motion itself, but is rather a *representation* that supports decisions, such as the decision where the robot should move to in order to manipulate. This goal position can then be given to a motion planner in order to find a trajectory that gets the robot to the goal position. In our system for instance, the navigation trajectory is determined by a Wavefront planner. Also, an ARPLACE is not a Reinforcement Learning policy (Sutton & Barto, 1998). A policy maps states to actions, whereas an ARPLACE maps (uncertain) states to expected probabilities of successfully executing a certain action. ARPLACES are thus *models of actions*, and not executable actions in their own right. This distinction will become most apparent in Section 5, in which ARPLACES are used by a transformational planner to detect and repair performance flaws in symbolic plans.

From this perspective, most similar to our work are ASYMOV (Cambon, Gravot, & Alami, 2004) and RL-TOPs (Ryan, 2002), in that they use symbolic planners to generate sequences of motion plans/reinforcement learning policies respectively. The specific contributions of this article are to enable the robot to learn grounded, probabilistic models of actions to support symbolic decision making, as well as using more flexible transformational planners that reason with these models. Our focus is thus more on grounding and improving the representations that enable symbolic planning, rather than the underlying actions that generate trajectories and/or the actual motion.

Okada, Kojima, Sagawa, Ichino, Sato, and Inaba (2006) also develop representations for place to enable symbolic planning, and they denote a good base placement for grasping a 'spot'. Different spots are hand-coded for different tasks, such as manipulating a faucet, a cupboard, and a trashcan. These symbolic representations of place are then used by a LISP-based motion planner to perform tool manipulation behavior. ARPLACE extends the concept of a 'spot' by learning it autonomously, grounding it in observed behavior, and providing a probabilistic representation of place. Berenson, Choset, and Kuffner (2008) address the issue of finding optimal start and goal configurations for manipulating objects in pick-and-place operations. They explicitly take the placement of the mobile base into account. As they are interested in the optimal start and goal configurations, instead of a probabilistic representation, this approach does not enable least-commitment planning. Diankov, Ratliff, Ferguson, Srinivasa, and Kuffner (2008) use a model of the reachable workspace of the robot arm to decide where the robot may stand to grasp an object and to focus the search. However, uncertainties in the robot's base position or the object's position are not considered, and thus cannot be compensated for. More recent work by Berenson, Srinivasa, and Kuffner (2009) addresses these issues, but still relies on an accurate model of the environment, and at a high computational cost. On the other hand, ARPLACE is a compact representation that is computed with negligible computational load, allowing for continuous updating.

Recently, similar methods to the ones presented in this article have been used to determine successful grasps, rather than base positions for grasping. For instance, Detry et al. (2009) determine a probability density function that represents the graspability of specific objects. This function is learned from samples of successful robot grasps, which are biased by observed human grasps. However, this approach does not take examples of failed grasps into account. As we shall see in Section 4, the distance between a failed and a successful grasp can be quite small, and can only be determined by taking failed grasps into account. Our classification boundaries in Section 3.2 are similar to Workspace Goal Regions, except that our boundaries refer to base positions, whereas Workspace Goal Regions refer to grasp positions (Berenson, Srinivasa, Ferguson, Romea, & Kuffner, 2009). Also, we generalize over these boundaries with a Point Distribution Model, and use it to generate a probabilistic concept of successful grasp positions.

Kuipers, Beeson, Modayil, and Provost (2006) present a bootstrapping approach that enables robots to develop high-level ontologies from low-level sensor data including distinctive states, places, objects, and actions. These high level states are used to choose trajectory-following control laws to move from one distinctive state to another. Our approach is exactly the other way around: given the manipulation and navigation skills of the robot (which are far too high-dimensional to learn with trajectory-following control laws), learn places from which these skills (e.g. grasping) can be executed successfully. Our focus is on action and affordance, not recognition and localization. For us, place means 'a cluster of locations from which I can execute my (grasping) skill successfully', whereas for Kuipers et al. it refers to a location that is perceptually distinct from others, and can therefore be well-recognized. Furthermore, their work has not yet considered the physical manipulation of objects, and how this relates to place.

Learning success models can be considered as probabilistic pre-condition learning. Most research in this field until now far focussed on learning symbolic predicates from symbolic examples (Clement, Durfee, & Barrett, 2007; Chang & Amir, 2006; Amir & Chang, 2008). These approaches have not been applied to robots, because the representations that are learned are not able to encapsulate the complex conditions that arise from robot dynamics and action parameterization. In robotics, the focus in pre-condition learning is therefore rather on grounding pre-conditions in robot experience. A more realistic domain is considered by Zettlemoyer, Pasula, and Kaelbling (2005), where a simulated gripper stacks objects in a blocks world. Here, the focus is on predicting possible outcomes of the actions in a completely observable, unambiguous description of the current state; our emphasis is rather on taking state estimation uncertainty into account. 'Dexter' learns sequences of manipulation skills such as searching and then grasping an object (Hart, Ou, Sweeney, & Grupen, 2006). Declarative knowledge such as the length of its arm is learned from experience. Learning success models has also been done in the context of robotic soccer, for instance learning the success rate of passing (Buck & Riedmiller, 2000), or approaching the ball (Stulp & Beetz, 2008). Our system extends these approaches by explicitly representing the regions in which successful instances were observed, and computing a *Generalized* Success Model for these regions.

An interesting line of research that shares some paradigms with ARPLACEs, such as learning the relation between objects and actions or building prediction models, are Object-Action Complexes (OACs). Geib, Mourao, Petrick, Pugeault, Steedman, Krüger, and Wörgötter (2006) and Pastor, Hoffmann, Asfour, and Schaal (2009) present OACs that can be used to integrate high-level artificial intelligence planning technology and continuous low-level robot control. The work stresses that, for a cognitive agent, objects and actions are inseparably intertwined and should therefore be paired in a single interface. By physically interacting with the world and applying machine learning techniques, OACs allow the acquisition of high-level action representations from low-level control representations. OACs are meant to generalize the principle of affordances (Gibson, 1977).

## 3. Learning a Generalized Success Model for ARPlace

In this section, we describe the implementation of the off-line phase depicted in Figure 3, in which a Generalized Success Model (GSM) is learned. The goal is to acquire the function $g$

$$P(Success|\mathbf{f}^{rob}, \mathbf{f}^{obj}) \quad = \quad g(\mathbf{f}^{rob}, \mathbf{f}^{obj}) \mapsto \{0, 1\} \tag{2}$$

which predicts the chance of a successful manipulation action, given the relative positions of the robot and the object, which are stored in the feature vectors $\mathbf{f}^{rob}$ and $\mathbf{f}^{obj}$ respectively. Note that during the off-line learning phase, these are *known* positions. Uncertainty in positions is taken into account in the on-line phase, as described in Section 4.

Performing mobile manipulation is a complex task that involves many hardware and software modules. An overview of these modules in our platform is described in Appendix A. This overview demonstrates the large number of modules required to implement a mobile manipulation platform. Many of these modules themselves are the results of years if not decades of research and development within companies, research groups, and open-source projects. The global behavior of the robot, e.g. whether it can grasp a cup from a certain base position, depends on *all* of these modules, and the interactions between them. In some cases, analytic models of certain modules are available (such as a Capability Map for the arm's workspace, Section 3.1). However, there is no general way of composing such models to acquire a global model of the system's behavior during task execution. Therefore, we rather *learn* this model from observed experience.

However, the component that computes ARPLACEs requires exactly such a global model to predict under which circumstances a manipulation action will fail or succeed. As attempting a theoretical analysis to model all foreseeable events and uncertainties about the world is at best tedious and error-prone, and at worst infeasible, we therefore use experience-based learning to acquire global models of the behavior. By doing so, the model is grounded in the observation of actual robot behavior.

The off-line learning phase consists of three steps: 1) repeatedly execute the action sequence and observe the result for $N$ different target object positions; 2) learn $N$ Support Vector Machines classifiers for $N$ specific cup positions 3) generalize over these $N$ classifiers with a Point Distribution Model.

### 3.1 Data Acquisition

The robot acquires experience by executing the following action sequence: 1) navigate to a specified base position at the table; 2) reach for the cup; 3) close the gripper; 4) lift the cup (Stulp, Fedrizzi, & Beetz, 2009a). In this action sequence, the task context is determined by the following parameters 1) the pose to which the robot navigates to[4]; 2) the

---

4. Note that the navigation planner is parameterized such that the robot is always directly facing the table. This is not a limitation of the planner, but rather a constraint that is added to make the behavior of pose of the target object on the table. After execution, the robot logs whether the object was successfully grasped or not. To efficiently acquire sufficient data, we perform the training experiments in the Gazebo simulator (Gerkey, Vaughan, & Howard, 2003). The robot is modeled accurately, and thus the simulator provides training data that is also valid for the real robot. Examples of a failed and successful grasp are depicted in Figure 4.

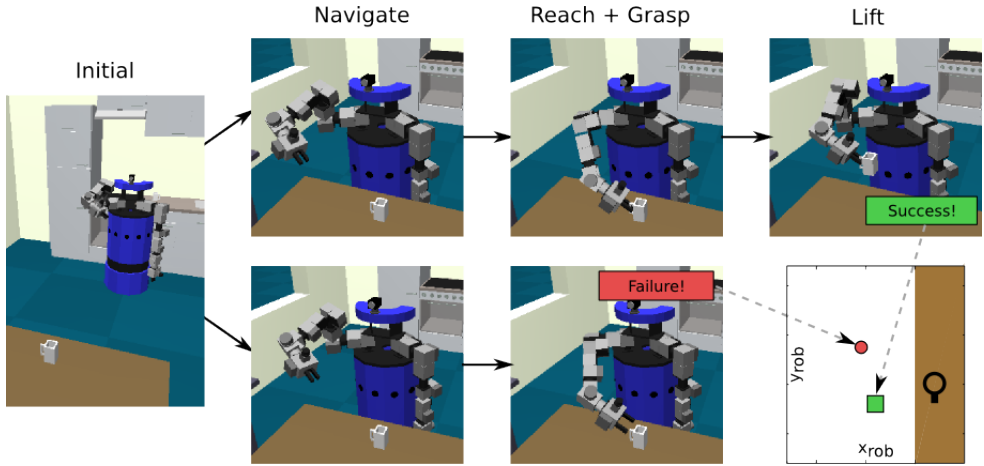

Figure 4: Two experiment runs with different samples for the robot position. The navigate-reach-grasp sequence in the upper row succeeds, but fails in the lower sequence.

The vector field controller we use to perform the reaching movement has proven to be robust in a wide range of the robot's workspace (Beetz et al., 2010). It also has a very low computational load, is easy to debug, and can quickly be adapted to novel objects. A disadvantage is that it occasionally gets stuck in local minima, after which the motion must be restarted. Our probabilistic motion planner for the arm does not suffer from local minima, but plan generation fails at the border of the workspace; even though the vector field controller is able to grasp there. Every planner and controller has its advantages and disadvantages, and there will always be sources of failure in the real world, especially for complex embodied agents. This article aims at modelling those failures through experience-based learning, and basing decisions on these models.

The feature space in which the data is collected is depicted in Figure 5. This coordinate system is relative to the table's edge, and the position of the cup on the table. This will enable us to apply the model that is learned from the data to different tables at different locations in the kitchen, in contrast to our previous work (Stulp, Fedrizzi, & Beetz, 2009b). From now on, we will refer to $\mathbf{f}^{obj} = [\Delta x^{obj} \Delta \psi^{obj}]$ as the *observable* task-relevant parameters, which the robot observes but cannot influence directly. Here $\Delta x^{obj}$ is the distance of the object to the table edge, and $\Delta \psi^{obj}$ is the angle between the object orientation and the normal that goes through the table edge and the object, as depicted in

---

the physical robot more predictable; this makes the robot more safe, which is required to operate the robot in human environments (cf. Figure 20). In principle, the methods in this paper could take this orientation into account.

Figure 5. $\mathbf{f}^{rob} = [\Delta x^{rob} \Delta y^{rob}]$ are the *controllable* action parameters, because the robot can use its navigation system to change them.

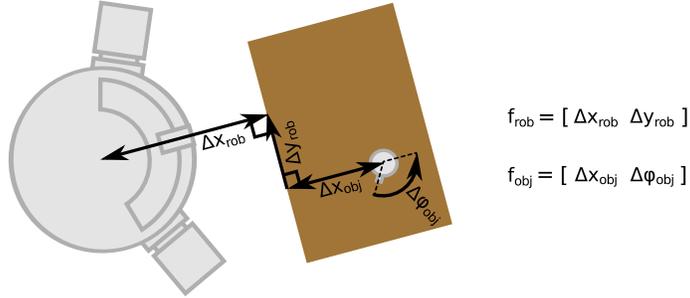

$$f_{rob} = [\; \Delta x_{rob} \;\; \Delta y_{rob} \;]$$

$$f_{obj} = [\; \Delta x_{obj} \;\; \Delta \varphi_{obj} \;]$$

Figure 5: Relative feature space used in the rest of this paper.

The robot gathers data for 16 target object poses, as depicted in Figure 6. The target object poses are listed in the matrix $\mathbf{F}^{obj}$. For a given target object position, we determine a rectangular area which is our generous estimation of the upper bound from which the robot can grasp the object. This rectangle is the same for all 16 object poses. Within this rectangle a uniform grid of almost 200 positions, which are stored in the matrix $\mathbf{F}^{rob}$, is defined. Figure 6 depicts the results of data gathering for these positions. Here, the markers represent the position of the robot base at the table. There are three types of markers, which represent the following classes:

### 3.1.1 Theoretically Unreachable (Light Round Markers)

The cup cannot be grasped from many of the positions in the bounding rectangle simply because the arm is not long enough. More formally, for these base positions, the kinematics of the arm are such that no inverse kinematic solution exists for having the end-effector at the position required to grasp the target object. We exploit analytic models of arm kinematics to filter out base positions in the bounding rectangle from which the cup is theoretically unreachable. The analytic model we use is a capability map, which is a compiled representation of the robot's kinematic workspace (Zacharias, Borst, & Hirzinger, 2007). Capability maps are usually used to answer the question: given the position of my base, which positions can I reach with my end-effector? In this article, we use the capability map to answer the inverse question: given the position of the target object (and therefore the desired position of my end-effector), from which base positions can I reach this end-effector position? In Figure 7, the answer to this question is visualized for a specific target object position. The depicted area is a theoretical kinematic upper bound on the base positions from which the robot can reach the target.

For each example base position in the bounding box, we use the capability map to determine if the target object is theoretically reachable. If not, the corresponding base position is labeled as a failure without executing the navigate-reach-grasp action sequence. This saves time when gathering data. Another obvious theoretical bound we implemented was that the robot's distance to the table should be at least as big as the robot's radius. Otherwise the robot would bump into the table. Again, we labeled such base positions as failures without executing them in order to save time.

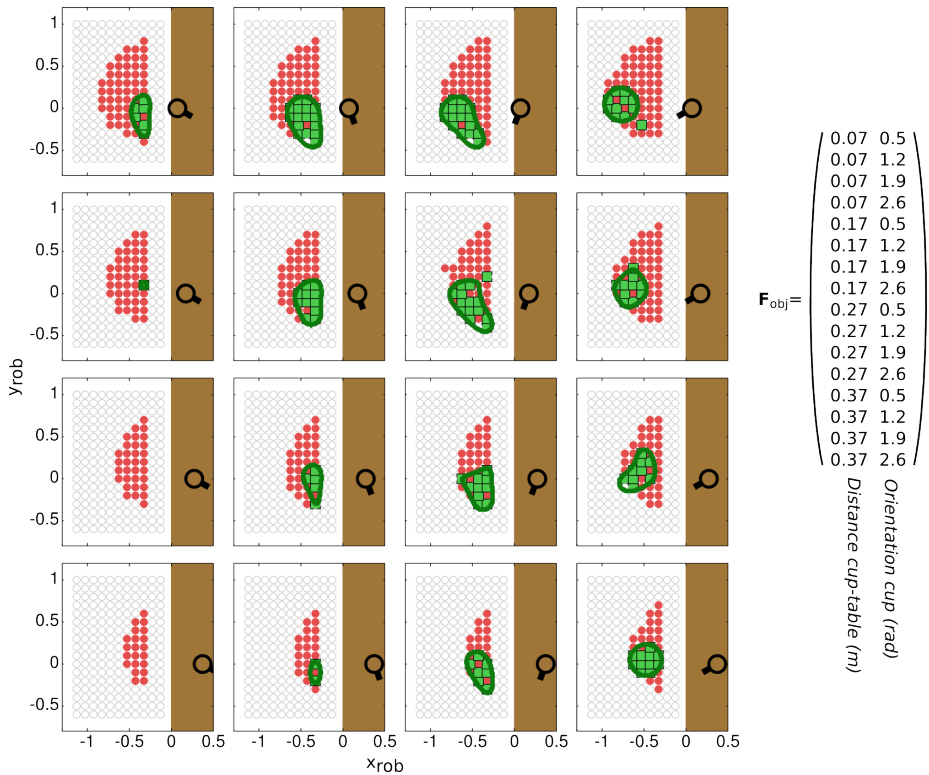

Figure 6: Results of data acquisition for 16 target object poses, listed in the matrix $\mathbf{F}^{obj}$. Markers correspond to the center of the robot base. Green squares and red circles represent successful and failed grasps respectively. Bright circles were not executed as a successful grasp is deemed theoretically impossible by capability maps. The dark green hulls are the classification boundaries (Section 3.2).

### 3.1.2 PRACTICALLY UNREACHABLE (RED FILLED ROUND MARKERS)

The capability map only considers the theoretical reachability of a position, given the kinematics of the robot's arm. It does not take self-collisions into account, or the constraints imposed by our vector-field controller for reaching, or the specific hardware of our gripper, and the way the gripper interacts with the target object. Red markers in Figure 6 represent base positions that the capability map deems possible, but that lead to a failure while performing the reaching motion. Some causes for failure are: 1) bumping into the table due to imprecision in the navigation routine 2) bumping into the cup before grasping it; 3) closing the gripper without the cup handle being in it; 4) the cup slipping from the gripper 5) the vector field controller getting caught in a local minimum.

One aim of this article is to demonstrate how such practical problems, that arise from the interaction of many hard- and software modules, are properly addressed by experience-based learning. Our approach is to use analytic models when they are available, but use experience-based learning when necessary. By interacting with the world, the robot observes its global behavior, and learns the difference between what is possible in theory and what works in practice.

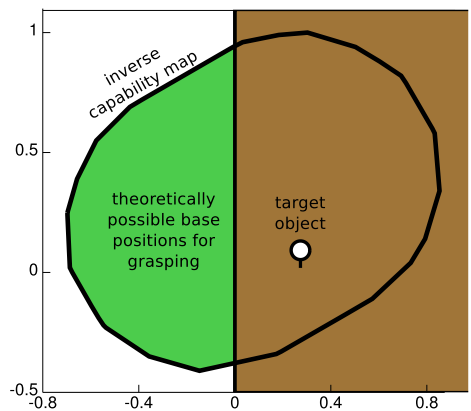

Figure 7: Inverse capability map for the right arm for a specific object position.

### 3.1.3 Reachable (Green Square Markers)

These are base positions from which the robot was able to successfully grasp the cup. The task execution is deemed successful when the cup is more than 10cm above the table when the action sequence is completed, because this can only be the case if the robot is holding it. We prefer such an empirical measure over for instance a force-closure measure, as the latter requires accurate models of the object, which we do not always have. Furthermore, it has been argued on theoretical grounds (Zheng & Qian, 2005), as well as demonstrated empirically (Morales, Chinellato, Fagg, & del Pobil, 2004), that force-closure grasps may not always lead to successful grasps in practice. Of course, force-closure may just as well be used as a measure of successful grasping; the methods described in this article do not depend upon this design choice.

The data acquisition yields a set of discrete robot and object positions, associated with the resulting outcome of the manipulation action, being a success or a failure:

$$P(Success|\{\mathbf{f}^{rob}\}_{i=1}^{M}, \{\mathbf{f}^{obj}\}_{j=1}^{N}) = b_{i,j}, \text{ with } b_{i,j} \in \{0, 1\} \tag{3}$$

In this article, the number of sampled object positions $N = 4 \times 4 = 16$ (i.e. the number of graphs in Figure 6), and the number of sampled robot positions $M = 11 \times 17 = 187$ (i.e. the number of data points per graph in Figure 6).

In the remainder of this section, we first generalize over the $M$ discrete robot positions by training Support Vector Machines (Section 3.2), and then generalize over the $N$ cup positions with Point Distribution Models (Section 3.3)

## 3.2 Generalization over Robot Positions

In this step, we generalize over the discrete robot positions, and acquire a compact boolean classifier that efficiently predicts whether manipulation will succeed:

$$P(Success|\mathbf{f}^{rob}, \{\mathbf{f}^{obj}\}_{j=1}^N) \quad = \quad g_{j=1\ldots N}(\mathbf{f}^{rob}) \mapsto \{0, 1\} \tag{4}$$

This generalization is implemented as follows. A separate classifier $g_{i=1\ldots N}$ is learned for each of the $N = 16$ object poses, i.e. one classifier for each of the 16 data sets depicted in Figure 4. To acquire these prediction models, we compute a classification boundary around the successful samples with Support Vector Machines (SVM), using the implementation by Sonnenburg et al. (2006), with a Gaussian kernel with $\sigma$=0.1 and cost parameter $C$=40.0. As successful grasps are rarer, we weight them twice as much as failed grasp attempts. Figure 6 depicts the resulting classification boundaries for different configurations of task-relevant parameters as dark-green boundaries. Manipulation is predicted to succeed if the robot's base position lies within a boundary for a given target object pose $\in \mathbf{F}^{obj}$. The accuracy of these learned classifiers is listed in Section 6.1.

### 3.3 Generalization over Object Positions

In the next step, we generalize over the discrete object positions:

$$P(Success|\mathbf{f}^{rob}, \mathbf{f}^{obj}) \quad = \quad g(\mathbf{f}^{rob}, \mathbf{f}^{obj}) \mapsto \{0, 1\} \tag{5}$$

We do so by determining a low-dimensional set of parameters that allows us to interpolate between the individual classification boundaries that the Support Vector Machines generate. This is done with a Point Distribution Model (PDM), which is an established method for modelling variations in medical images and faces (Cootes et al., 1995; Wimmer, Stulp, Pietzsch, & Radig, 2008). The result is one compact model that incorporates the individual boundaries, and is able to interpolate to make predictions for target object poses not observed during training.

As input a PDM requires $n$ points that are distributed over a contour. How these landmarks are distributed is described in Appendix B. Given the landmarks on the classification boundaries, we compute a PDM. Although PDMs are most well-known for their use in computer vision (Cootes et al., 1995; Wimmer et al., 2008), we use the notation by Roduit, Martinoli, and Jacot (2007), who focus on robotic applications. First, the 16 boundaries of 20 2D points are merged into one 40x16 matrix $\mathbf{H}$, where the columns are the concatenation of the $\Delta x^{rob}$ and $\Delta y^{rob}$ coordinates of the 20 landmarks along the classification boundary. Each column thus represents one boundary. The next step is to compute $\mathbf{P}$, which is the matrix of eigenvectors of the covariance matrix of $\mathbf{H}$. $\mathbf{P}$ represents the principal modes of variation. Given $\overline{\mathbf{H}}$ and $\mathbf{P}$, we can decompose each boundary $\mathbf{h}_{1..16}$ in the set into the mean boundary and a linear combination of the columns of $\mathbf{P}$ as follows $\mathbf{h}_k = \overline{\mathbf{H}} + \mathbf{P} \cdot \mathbf{b}_k$. Here, $\mathbf{b}_k$ is the so-called deformation mode of the $k^{th}$ boundary. This is the Point Distribution Model. To get an intuition for what the PDM represents, the first three deformation modes are depicted in Figure 8, where the values of the first, second and third deformation modes (columns 1, 2, 3 of $\mathbf{B}$) are varied between their maximal and minimal value, whilst the other deformation modes are set to 0.

The eigenvalues of the covariance matrix of $\mathbf{H}$ indicate that the first 2 components already contain 96% of the deformation energy. For reasons of compactness and to achieve

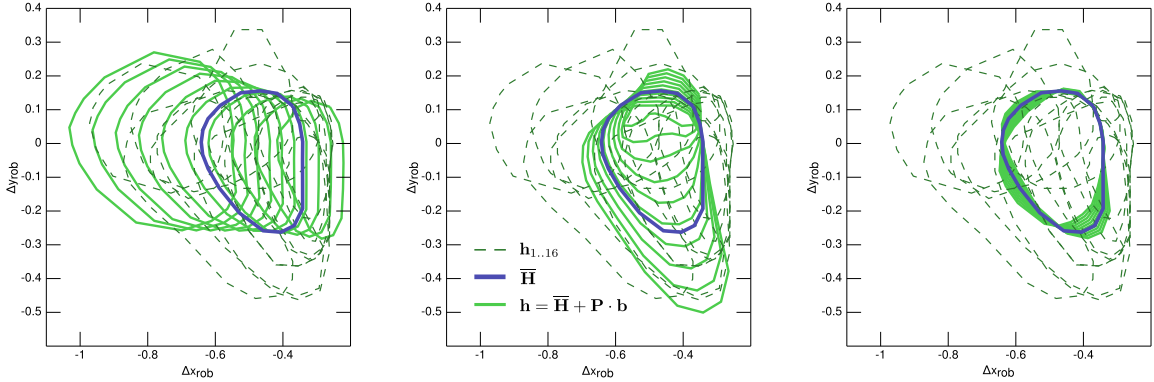

Figure 8: The first 3 deformation modes of the Point Distribution Model (in **B**).

better generalization, we use only the first 2 deformation modes, without losing much accuracy.

$$\{\mathbf{f}_j^{obj}, \quad g_j(\mathbf{f}^{rob}) = \texttt{inboundary}(\mathbf{f}^{rob}, \mathbf{h}_j)\}_{j=1}^N \qquad N \text{ Support Vector Machines} \qquad (6)$$

$$\{\mathbf{f}_j^{obj}, \quad g_j(\mathbf{f}^{rob}) = \texttt{inboundary}(\mathbf{f}^{rob}, \overline{\mathbf{H}} + \mathbf{P} \cdot \mathbf{b}_j)\}_{j=1}^N \qquad \text{Point Distribution Model} \qquad (7)$$

$$g(\mathbf{f}^{rob}, \mathbf{f}^{obj}) = \texttt{inboundary}(\mathbf{f}^{rob}, \overline{\mathbf{H}} + \mathbf{P} \cdot b(\mathbf{f}^{obj})) \qquad \text{Regression between } \mathbf{f}_j^{obj} \text{ and } \mathbf{b}_j \qquad (8)$$

The PDM has several advantages: 1) instead of having to store $N = 16$ classification boundaries $\mathbf{h}_j$ with each 20 2D points to capture the variation in classification hulls due to different target object positions, we only store $N = 16$ deformation modes with 2 degrees of freedom each. This greatly reduces the dimensionality; 2) the 2 degrees of freedom in $\mathbf{b}$ can be used to interpolate in a principled way between the computed classification boundaries $\mathbf{h}_j$, to generate boundaries for object positions that were not observed during learning; 3) a simple regression between the two degrees of freedom of the PDM $\mathbf{b}$ and the position $\mathbf{f}^{obj}$ is feasible, so that the object position can be related directly to the shape of the classification boundary. This regression is explained in the next section.

### 3.4 Relation to Task-Relevant Parameters

In this step, we acquire a function $b$, that computes the appropriate deformation modes $\mathbf{b}$ for a given object position $\mathbf{f}^{obj}$. To do so, we compute a regression between the matrix of deformation modes of the specific object positions $\mathbf{B}$, and the 16 object positions themselves in $\mathbf{F}^{obj}$, as depicted in Figure 6. We found that a simple second order polynomial regression model suffices to compute the regression, as it yields high coefficients of determination of $R^2 = 0.99$ and $R^2 = 0.96$ for the first and second deformation modes respectively. The coefficients of the polynomial model are stored in two 3x3 upper triangular matrices $\mathbf{W}_1$ and $\mathbf{W}_2$, such that $\mathbf{B} \approx [\ diag([\mathbf{T}\ \mathbf{1}] \cdot \mathbf{W}_1 \cdot [\mathbf{F}^{obj}\ \mathbf{1}]^T)\ diag([\mathbf{T}\ \mathbf{1}] \cdot \mathbf{W}_2 \cdot [\mathbf{T}\ \mathbf{1}]^T\ ]$

The Generalized Success Model now consists of 1) $\overline{\mathbf{H}}$, the mean of the classification boundaries computed with the SVM; 2) $\mathbf{P}$, the principal modes of variation of these clas-sification boundaries; 3) $\mathbf{W_{1,2}}$, the mapping from task-relevant parameters to deformation modes.

Let us now summarize how the Generalized Success Model is used to predict successful manipulation behavior:

1. The Generalized Success Model takes the (observed) relative position of the object on the table $\mathbf{f}_{cur}^{obj} = [\Delta x_{cur}^{obj}\ \Delta \psi_{cur}^{obj}]$ as input (Figure 5).

2. The appropriate deformation values for the given object position are computed with $\mathbf{b_{cur}} = b(\mathbf{f}_{cur}^{obj}) = [\ \mathbf{q} \cdot \mathbf{W_1} \cdot \mathbf{q}^T\quad \mathbf{q} \cdot \mathbf{W_2} \cdot \mathbf{q}^T\ ]$, where $\mathbf{q} = [\mathbf{f}_{cur}^{obj}\ 1] = [\Delta x_{cur}^{obj}\ \Delta \psi_{cur}^{obj}\ 1]$ (Section 3.4).

3. The boundary is computed with $\mathbf{h_{cur}} = \overline{\overline{\mathbf{H}}} + \mathbf{P} \cdot \mathbf{b_{cur}}$ (Section 3.3).

4. If the relative robot base center $\mathbf{f}^{rob} = [\Delta x^{rob}\ \Delta y^{rob}]$ is within boundary $\mathbf{h_{cur}}$, the model predicts that the robot will be able to successfully grasp the object at position $\mathbf{f}_{cur}^{obj} = [\Delta x_{cur}^{obj}\ \Delta \psi_{cur}^{obj}]$.

Note that these steps only involve simple multiplications and additions of small matrices, and thus can be performed very efficiently[5]. The reason for this efficiency lies in the fact that we directly relate task-relevant parameters, such as the position of the cup on the table, to predictions about the *global* behavior of the robot, such as whether the manipulation action will succeed or not. The on-line efficiency is made possible by *experience-based learning*, where the wealth of information in the observation of global behavior is compiled into a compact model off-line. This approach adheres to the proposed strategy of *"learning task-relevant features that map to actions, instead of attempting to reconstruct a detailed model of the world with which to plan actions"* (Kemp, Edsinger, & Torres-Jara, 2007).

In summary, from observed behavior outcomes, we have learned the mapping in Equation 2 (which we repeat in Equation 9), which maps continuous robot and target object positions to a boolean prediction about the success of an action:

$$P(Success|\mathbf{f}^{rob}, \mathbf{f}^{obj}) = g(\mathbf{f}^{rob}, \mathbf{f}^{obj}) \mapsto \{0, 1\} \tag{9}$$

This mapping can be used to predict if the current base position of the robot will lead to successful manipulation, but also to determine appropriate base positions to navigate to.

Equation 9 assumes that the true values of the robot and target object positions are known to the robot. In Section 4, we discuss how uncertainties in the estimates of these positions is taken into account during task execution.

## 3.5 Generality of Generalized Success Model

Before explaining how the Generalized Success Model is used to generate ARPLACEs on-line during task-execution in Section 4, we discuss some of the generalization properties and limitations of the Generalized Success Model. To do so, we must distinguish between the *general applicability of our approach* to different robots, objects and domains, and the *specificity* of the model to these factors *once it has been learned*. This essentially holds for any data-driven approach: the model can in principle be learned for any data, independent of the robot system that generates these data, but once learned, will be specific to the data generated by that robot system, and thus specific to the robot system itself. For all practical purposes, we assume that the domain and robot hardware remain fixed, so learning a domain- and robot-specific model is not a grave limitation.

---

5. As an indication, with the model that is described in this paper all four steps take 0.2ms on a 2.2GHz machine in our Matlab implementation.

### 3.5.1 Generalization over Object Poses

The learned model generalizes over different object poses, as the relative object pose on the table $\mathbf{f}^{obj} = [\Delta x^{obj} \ \Delta \psi^{obj}]$ is part of the feature space with which the Generalized Success Model is parameterized (see step 1. in calling a GSM in Section 3.4). The 'Generalized' actually refers to this capability of generalizing over Success Models for specific object poses.

### 3.5.2 Grasp-Specific ARPlaces

By being specific to the object, a lot of data would be required to learn an ARPlace for each object the robot should manipulate. In practice however, we found that only a few grasps suffice to grasp most everyday objects in kitchen environments with the real robot platform (Maldonado, Klank, & Beetz, 2010). In particular, this approach required only 2 grasps (one from the top and one from the side) to achieve 47 successful grasps out of 51 attempts with 14 everyday kitchen objects. Therefore, we propose to use *grasp-specific* ARPlaces, rather than object-specific ARPlaces. We have learned Generalized Success Models for both grasps, which are depicted in Figure 9.

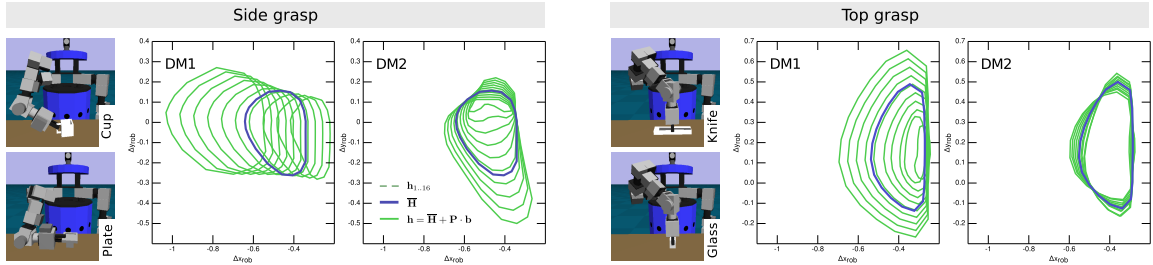

Figure 9: The two Point Distribution Models for the side and top grasp. Examples of objects that can be manipulated with these grasps are depicted.

The two deformation modes for the Point Distribution Model depicted in Figure 9 already contain 99% of the deformation energy, which is even more than for the side grasps. This is because the success of the side grasp is relatively independent of the orientation of the object, as the robot does not need to reach around the object. This also leads to more symmetric classification boundaries for the top grasp, as can be seen in Figure 9.

In summary, only two Generalized Success Models must be learned for two different grasps, as these two grasps suffice to grasp the 14 everyday kitchen objects that were tested with the real robot by Maldonado et al. (2010). In the rest of this article, we will focus on the side grasp; ARPLACEs for top grasps are presented by Fedrizzi (2010).

## 4. Computing Action-Related Places

In the previous section, we demonstrated how the Generalized Success Model is learned from observed experience for a variety of task parameterizations. The resulting function maps known robot and object positions to a prediction whether the action execution will succeed or feel.

In this section, we describe how ARPLACEs for manipulation are computed on-line for specific task contexts. As depicted in Figure 3, this module takes the Generalized Success Model and the estimated robot pose and target object pose as input, and returns an ARPLACE such as depicted in Figure 1.

### 4.1 Taking Object Position Uncertainty into Account

In Equation 9, the prediction whether a manipulation action succeeds or fails is based on *known* robot and target object positions. However, during task execution, the robot only has estimates of these positions, with varying levels of uncertainty. These uncertainties must be taken into account when predicting the outcome, as a manipulation action that is predicted to succeed might well fail if the target object is not at the position where the robot expects it to be. Given the Generalized Success Model in Equation 9, the goal of this section is therefore to compute the mapping

Generalized Success Model                                ARPLACE (with object uncertainty)

$$P(Succ|\mathbf{f}^{rob}, \mathbf{f}^{obj}) \mapsto \{0,1\} - \boxed{\text{Monte Carlo}} \rightarrow \ \{ \ P(Succ|\mathbf{f}_k^{rob}, \langle \hat{\mathbf{f}}^{obj}, \mathbf{\Sigma}^{obj} \rangle) \ \}_{k=1}^{K} \mapsto [0,1] \tag{10}$$

which takes *estimates* of the target object position, and returns a *continuous* probability value, rather than a discrete $\{0,1\}$ probability value as in Equation 9. In our belief state, the uncertainties in object positions are modelled as a Gaussian distribution with mean $\hat{\mathbf{f}}^{obj}$ and covariance matrix $\mathbf{\Sigma}^{obj}$.

On the robot platform described in Appendix A, $\hat{\mathbf{f}}^{obj}$ and $\mathbf{\Sigma}^{obj}$ are obtained from a vision-based object localization module (Klank, Zia, & Beetz, 2009). Typical values along the diagonal of the 6x6 covariance matrix are: $\sigma_{x,x}^2 = 0.05$, $\sigma_{y,y}^2 = 0.03$, $\sigma_{z,z}^2 = 0.07$, $\sigma_{yaw,yaw}^2 = 0.8$, $\sigma_{pitch,pitch}^2 = 0.06$, $\sigma_{roll,roll}^2 = 0.06$. The uncertainties in position are specified in meters and the angular uncertainties are specified in radians. The estimation of the object position is quite accurate, but our vision system has problems to detect the handle, which is important for estimating the orientation ($yaw$) of the cup. Due to the constraints enforced by our assumption that the cup is standing upright on the table, the uncertainty in $z$, *pitch* and *roll* is set to 0. The remaining 3x3 covariance matrix is mapped to the relative feature space, which yields $\mathbf{\Sigma}^{obj}$.

At the end of Section 3.4, we demonstrated how a classification boundary $\mathbf{h}_{new}$ is reconstructed, given known task relevant parameters $\mathbf{f}_{new}^{obj} = [\,\Delta x_{new}^{obj}\ \Delta \psi_{new}^{obj}\,]$. Because of the uncertainty in $\hat{\mathbf{f}}_{new}^{obj}$, it does not suffice to compute only one classification boundary given the most probable position of the cup as the ARPlace from which to grasp. This might lead to a failure if the cup is not at the position where it was expected. Therefore, we use a Monte-Carlo simulation to generate a whole set of classification boundaries. This is done by taking 100 samples from the Gaussian distribution of the object position, given its mean position and associated covariance matrix. This yields a matrix of task-relevant parameters $\mathbf{F}^{obj}_{s=1...100} = [\boldsymbol{\Delta x^{obj}}_s \ \boldsymbol{\Delta \psi^{obj}}_s]$. The corresponding classification boundaries are computed for the samples with $\mathbf{h}_s = \overline{\overline{\mathbf{H}}} + \mathbf{P} \cdot b(\mathbf{f}^{obj}_s))$ from Equation 8. In Figure 10(a), 20 out of the 100 boundaries are depicted. These were generated with the task-relevant parameters $\hat{\mathbf{f}}^{obj} = [\ \Delta x^{obj} \ \Delta \psi^{obj}\ ] = [\ 0.2 \ 1.5\ ]$ and $\boldsymbol{\Sigma}^{obj} = \begin{bmatrix} \sigma^2_{\Delta x^{obj} \Delta x^{obj}} & \sigma^2_{\Delta x^{obj} \Delta \psi^{obj}} \\ \sigma^2_{\Delta \psi^{obj} \Delta x^{obj}} & \sigma^2_{\Delta \psi^{obj} \Delta \psi^{obj}} \end{bmatrix} = \begin{bmatrix} 0.03^2 & 0 \\ 0 & 0.30^2 \end{bmatrix}$.

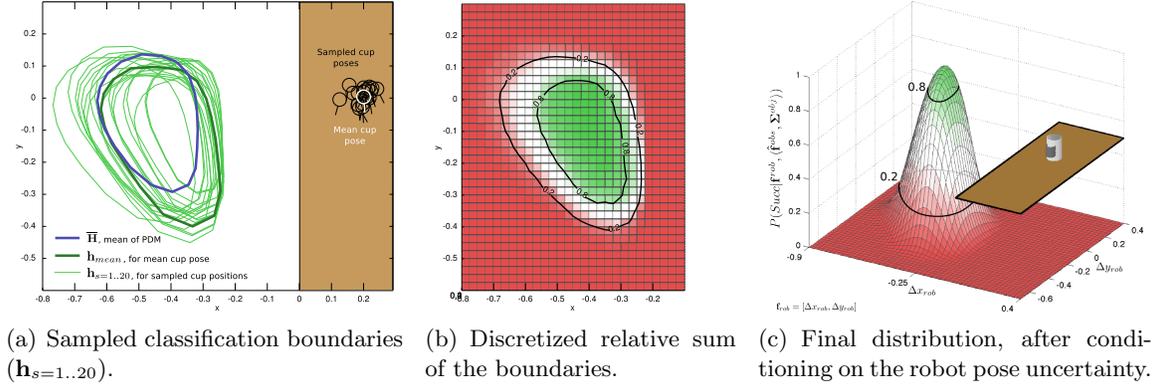

(a) Sampled classification boundaries ($\mathbf{h}_{s=1..20}$).

(b) Discretized relative sum of the boundaries.

(c) Final distribution, after conditioning on the robot pose uncertainty.

Figure 10: Monte-Carlo simulation of classification boundaries to compute ARPlace.

As described in Appendix C, $y$ is 0 by definition, as $F_{\text{GSM}}$ is defined relative to the cup's position along the table's edge. That is why the uncertainty in $y$, described by $\sigma^2_{y,y}$, leads to an uncertainty in the origin of $F_{\text{GSM}}$. Therefore, when sampling from the task-relevant parameters, we also sample values of $y$, and translate $F_{\text{GSM}}$ accordingly. Uncertainty in $y$ does not influence the *shape* of the classification boundary through the PDM, it simply translates the classification boundary along the table's edge. This sampling of $y$ has actually already been done in Figure 10(a), where $\sigma_{\Delta y^{obj} \Delta y^{obj}} = 0.03$. So in fact,

$$\boldsymbol{\Sigma}^{obj} = \begin{bmatrix} \sigma^2_{\Delta x^{obj} \Delta x^{obj}} & \sigma^2_{\Delta x^{obj} \Delta y^{obj}} & \sigma^2_{\Delta x^{obj} \Delta \psi^{obj}} \\ \sigma^2_{\Delta y^{obj} \Delta x^{obj}} & \sigma^2_{\Delta y^{obj} \Delta y^{obj}} & \sigma^2_{\Delta y^{obj} \Delta \psi^{obj}} \\ \sigma^2_{\Delta \psi^{obj} \Delta x^{obj}} & \sigma^2_{\Delta \psi^{obj} \Delta y^{obj}} & \sigma^2_{\Delta \psi^{obj} \Delta \psi^{obj}} \end{bmatrix} = \begin{bmatrix} 0.03^2 & 0 & 0 \\ 0 & 0.03^2 & 0 \\ 0 & 0 & 0.30^2 \end{bmatrix}$$

After having computed the sampled classification boundaries, we then generate a discrete grid of $2.5 \times 2.5 cm$ cells, which represent the discrete robot positions $\{\mathbf{f}^{rob}_k\}^K_{k=1}$ in Equation 1. For each cell, the number of classification boundaries that classify each cell as a success is counted. We are thus computing a histogram of predicted successful grasps. Dividing the result by the overall number of boundaries yields the probability that grasping the cup will succeed from this position. The corresponding distribution, which takes the uncertainty of the cup position into account, is depicted in Figure 10(b).

It is interesting to note the steep decline on the right side of the distribution near the table, where the probability of a successful grasp drops from 0.8 to 0.2 in about 5cm. This is intuitive, as the table is located on the right side, and the robot bumps into the table when moving to the sampled initial position, leading to an unsuccessful navigate-reach-grasp sequence. Therefore, none of the 16 boundaries contain the area that is close to the table, and the variation in **P** on the right side of the PDM is low. Variations in **B** do not have a large effect on this boundary, as can be seen in Figure 10(b). When summing over the sampled boundaries, this leads to a steep decline in success probability.

Note that an ARPLACE is not a normalized probability distribution (which sums to 1), but rather a probability *mapping*, in which each element (discrete grid cell) is a probability distribution itself. Thus the sum of probabilities in each grid cell *is* 1, i.e. $P(Succ) + P(\neg Succ) = 1$.

## 4.2 Taking Robot Position Uncertainty into Account

The robot not only has uncertainty about the position of the target object, but also about its own position. This uncertainty must also be taken into account in ARPLACE. For instance, although any position near to the left of the steep incline in Figure 10(b) is predicted to be successful, they might still fail if the robot is actually more to the right than expected. Therefore, we condition the probabilities in Figure 10(b) on the robot actually being at a certain grid cell $(\Delta x^{rob}, \Delta y^{rob})$ given its position estimate $(\hat{\Delta x}^{rob}, \hat{\Delta y}^{rob})[6]$, and acquire the final ARPLACE mapping as:

ARPLACE prob. mapping, Figure 10(c).

$$P(Success | \langle \hat{\mathbf{f}}^{rob}, \mathbf{\Sigma}^{rob} \rangle, \langle \hat{\mathbf{f}}^{obj}, \mathbf{\Sigma}^{obj} \rangle) =$$
$$P(Success | \mathbf{f}^{rob}, \langle \hat{\mathbf{f}}^{obj}, \mathbf{\Sigma}^{obj} \rangle) \cdot \quad P(\mathbf{f}^{rob} | \langle \hat{\mathbf{f}}^{rob}, \mathbf{\Sigma}^{rob} \rangle) \tag{11}$$

Prob. mapping Equation 10, Figure 10(b).    Prob. distribution robot uncertainty (Gaussian).

In this equation, $\langle \hat{\mathbf{f}}^{rob}, \mathbf{\Sigma}^{rob} \rangle$ can be interpreted in two ways. First of all, it can represent the actual estimate of the robot's position at the current time. In this case, $P(Success | \ldots)$ predicts the probability of success when manipulation from the current position. However, it can also be interpreted as possible goal positions the robot could navigate to in order to perform the navigation, i.e. $\langle \hat{\mathbf{f}}^{rob}_{goal}, \mathbf{\Sigma}^{rob}_{goal} \rangle$, as we do throughout this paper. In doing so, we make the assumption that the future position uncertainty $\mathbf{\Sigma}^{rob}_{goal}$ at the goal position $\hat{\mathbf{f}}^{rob}_{goal}$ is the same as it is currently, i.e. $\mathbf{\Sigma}^{rob}_{goal} = \mathbf{\Sigma}^{rob}$. We believe this is a fair assumption because; 1) it is more realistic than assuming $\mathbf{\Sigma}^{rob}_{goal} = \mathbf{0}$; 2) as the robot approaches the navigation goal, it is continually updating $\mathbf{\Sigma}^{rob}$, and thus $P(Success | \ldots)$. Once it has reached the goal, $\mathbf{\Sigma}^{rob}_{goal}$ will be equivalent to $\mathbf{\Sigma}^{rob}$.

## 4.3 Refining ARPlace On-line

In summary, ARPLACEs are computed on-line with a learned Generalized Success Model, given the task-relevant parameters of the current task context, which includes uncertainties in the poses of the robot and target object. This yields a probability mapping that maps robot base positions to the probability that grasping the target object will succeed.

---

6. Since the navigation planner is parameterized such that the robot always faces the table (cf. Section 3.1), we have ignored the orientation of the robot in computing the GSM. Note that we therefore also ignore the *uncertainty* in this parameter here, and ARPLACEs do not take it into account. We expect that the improved robustness (evaluated in Section 6.2) could be further improved by taking (the uncertainty) in this parameter into account.

Learning the Generalized Success Model is a costly step which involves extensive data collection, and thus is performed off-line. Once learned, this model is very compact, and is used to efficiently compute ARPlaces on-line[7] Therefore, ARPlaces can be updated as the execution of the task progresses, and can incorporate new knowledge about task-relevant parameters or changes in the environment. Figure 11 depicts how the ARPlace probability mapping is affected as new knowledge about task-relevant parameters comes in. The first row demonstrates how more accurate knowledge about the target object's position (lower uncertainty, e.g. lower $\sigma_{\Delta x^{obj}\Delta x^{obj}}$) leads to a more focussed ARPlace with higher overall probabilities, and a higher mode. The second and third row depict similar effects when estimates of the target object's position and orientation change. This figure serves two purposes: it gives the reader a visual intuition of the effects of several task-relevant parameters on the shape of the ARPlace, and it demonstrates how the robot's internal ARPlace representation might change as new (more accurate) information about the target object pose comes in.

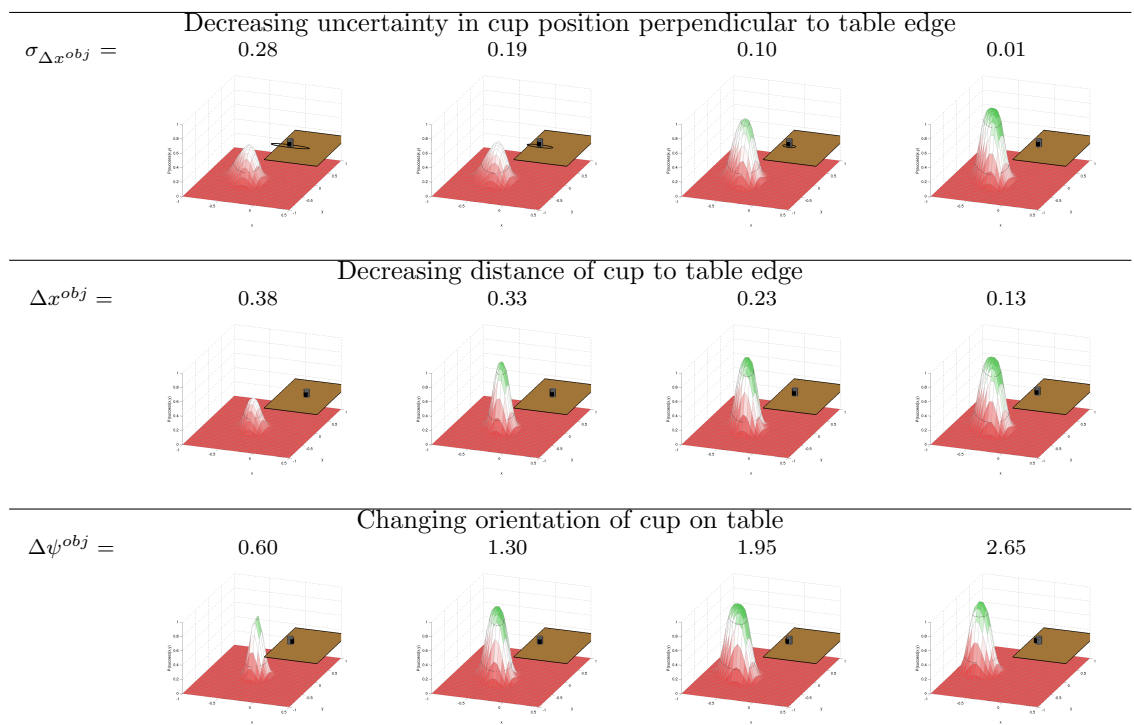

Figure 11: These images demonstrate how varying certain task-relevant parameters affects the shape of the ARPlace distribution.

The decision whether a certain probability of success suffices to execute the manipulation action critically depends on the domain and task. Failing to grasp a full glass of wine has more grave consequences than failing to grasp a tennis ball. In general, ARPLACE provides a representation which enables high-level planners to make rational decisions about such scenarios, but does not specify how such decisions should be made, or what the minimal success probability should be in order to perform the task. In Section 5 we present the use of ARPLACE in a concrete scenario.

---

7. As an indication, it takes on average 110ms on a 2.2GHz machine in our Matlab implementation to perform the steps in Section 4.1 and 4.2.

## 4.4 Generality of ARPlaces

In Section 3.5, we discussed the generality of *learning* the Generalized Success Model, and the specificity of the model with respect to the robot and its skills, once the model has been learned off-line. In this section, we demonstrate the generality and flexibility of the ARPLACE representation, which is generated on-line using the Generalized Success Model. We also present various ways in which ARPLACEs can be extended, and lay the groundwork for Section 5, which explains how ARPLACEs are used in the context of a high-level transformational planner.

### 4.4.1 MERGING ARPLACES FOR MULTIPLE ACTIONS

ARPLACEs for multiple actions can be composed by intersecting them. Assume we have computed ARPLACEs for two different actions ($a_1$ and $a_2$). If the success probabilities of the ARPLACEs is independent, we can compute the ARPLACE for executing both actions in parallel by multiplying the probabilities of the ARPLACEs for action $a_1$ and $a_2$.

In the first two graphs of Figure 12 for instance, the ARPLACEs for grasping a cup with the left and right gripper are depicted. With a piecewise multiplication of the probabilities, we acquire the merged ARPLACE, depicted in the right graph. The robot can use this merged ARPLACE to determine with which probability it can use the left and right gripper to grasp both cups from one base position (Fedrizzi, Moesenlechner, Stulp, & Beetz, 2009). Another similar application is merging the ARPLACEs for two cup positions, grasped with the same gripper. This ARPLACE represents the probability of being able to grasp a cup from one position, and placing it on the other position, without moving the base. Such compositions would be impossible if the robot commits itself to specific positions in advance.

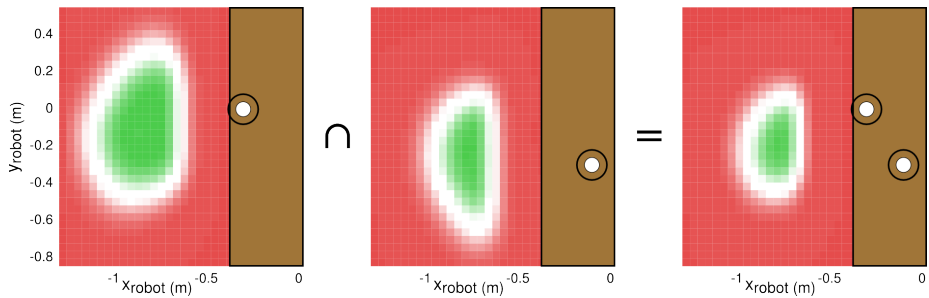

Figure 12: Left distribution: grasp cup with left gripper. Center distribution: grasp cup with right gripper. Right distribution (element-wise product of the other two distributions): Grasp both cups with left/right gripper from one base position.

As navigating to only one position to grasp two cups is much more efficient than navigating to two positions, we have implemented this decision as a transformation rule in the Reactive Planning Language (McDermott, 1991), which is described in detail in Section 5.

### 4.4.2 Different Supporting Planes

Defining the feature space of the Generalized Success Model relative to the table's edge allows the robot to compute ARPlaces for more general table shapes than the one presented so far. This is done by determining an ARPlace for each of the straight edges of a table, and computing the union of these individual ARPlaces. An example is depicted in Figure 13.

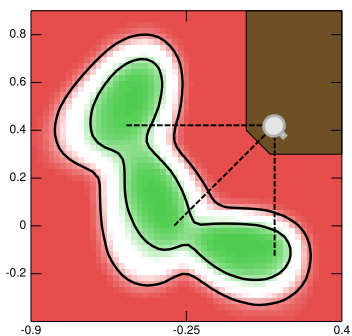

Figure 13: An ARPlace for a more complex table shape.

### 4.4.3 Different Uncertainty Distributions

In this article, the uncertainty in the position of the robot and target objects is modelled by a multi-variate Gaussian distribution. This is not because our approach expects such a distribution, but because this is how our state estimation systems represent uncertainty. In Section 4.1, we described how specific target object positions are sampled from this distribution in a Monte Carlo simulation. In general, our method applies to any distribution from which such a sampling can be done. These distributions need not be Gaussian, and might well be multi-modal or even non-parametric. For a particle filter for instance, each particle could directly be used as a sample to compute the classification boundaries as in Figure 10(a).

### 4.4.4 Applicability to Other Domains

We demonstrate the generality of ARPlaces by briefly showing how an ARPlace is able to represent a task-relevant place for a very different task and domain: approaching the ball in robotic soccer. This task frequently fails because the robot bumps into the ball before achieving the desired position at the ball. In Figure 14(a), examples of a successful (S) and failed (F) attempt are depicted. Here, the robot should approach the ball from the top. Our goal is to acquire an ARPlace that maps the robot's position on the field to the predicted probability that it will successfully approach the ball.

The procedure for learning an ARPLACE is equivalent to that in the mobile manipulation domain: 1) gather data and log successful and failed episodes (Figure 14(a)); 2) learn classification boundaries and a generalized success model from these data; 3) generate ARPLACES for specific task contexts (Figure 14(b)). This example demonstrates that the ARPLACE approach is not limited to mobile manipulation, but generalizes to other actions and domains.

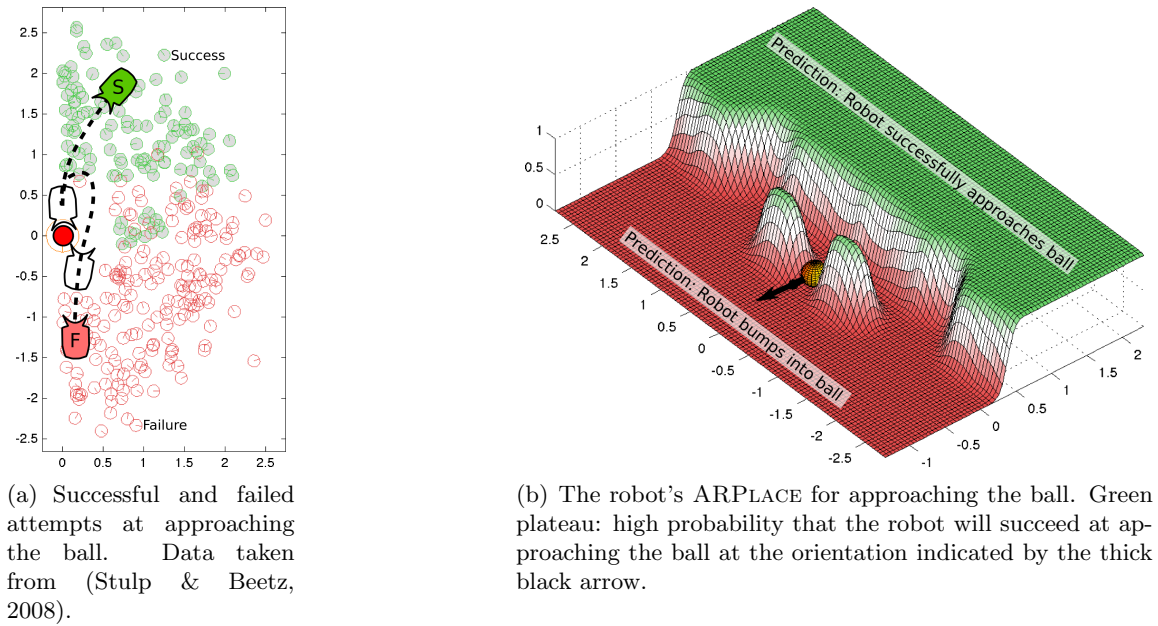

(a) Successful and failed attempts at approaching the ball. Data taken from (Stulp & Beetz, 2008).

(b) The robot's ARPLACE for approaching the ball. Green plateau: high probability that the robot will succeed at approaching the ball at the orientation indicated by the thick black arrow.

Figure 14: An ARPLACE for the robot soccer domain.

Note the two 'bumps' to the left and right of the ball. It is not intuitively clear why the robot should succeed in approaching the ball from these locations, but not the surrounding ones. We assume that it depends on the particular morphology of the robot, and the controller used to approach the ball; both are described by Stulp and Beetz (2008). One of the main advantages of using an approach based on learning is that our assumptions and intuitions do not play a role in acquiring the model. Whatever the reason may be, these successful approaches are obvious in the observed data (Figure 14(a)), and hence the ARPLACE represents them.

### 4.4.5 USING MORE GENERAL COST FUNCTIONS

In this article, the probability of success is considered the only utility relevant to determining an appropriate base position. But in principle, ARPLACE is able to represent any kind of utility or cost, an example of which is given in Figure 15. Here, the task of the robot is to collect one of the two cups on the table. The probabilistic ARPLACES for the two cups are depicted in the left graph. Given these parameters, the chance of success is 0.99 for both of the cups, so there is no reason to prefer fetching one over the other. However, cup B is much closer to the robot, and therefore it would be more efficient to collect cup B. This preference can be expressed with an ARPLACE. First, we compute the distance of the robot to each of the grid cells of the probabilistic ARPLACE, as depicted in the center graph. Finally, we merge the probability $P$ and distance $d$ into one cost $u$, with $u = (1 - P)5 + d/0.3$. This expresses that it takes on average 5 seconds to reposition the robot for another grasp attempt in case of a failure, and that the average navigation speed is $0.3m/s$. This cost thus expresses the expected time the overall task will take[8]

As depicted in Figure 15, the mode of the ARPLACE of the cup that is closer to the robot is now higher, reflecting the fact that we prefer the robot to fetch cups that are closer. For an in-depth discussion of utility-based ARPLACEs, and how they affect the behavior of the robot, we refer to Fedrizzi (2010).

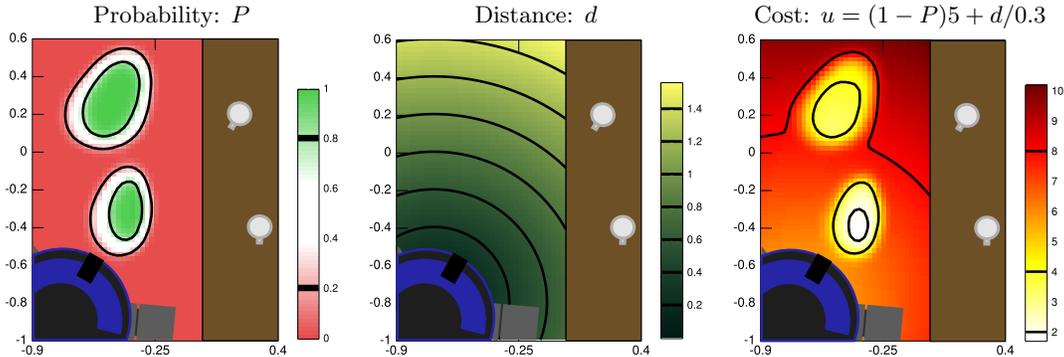

Figure 15: Example of a more general cost-based ARPLACE (right), including both the probability of success (left) and distance to the robot (center). By including the distance as part of the cost, the mode of the cost-based ARPLACE for the closer cup is higher than for the more distant cup.

# 5. Transformational Planning with ARPlace

So far, we have described how ARPLACEs are generated on-line by using the learned Generalized Success Model. The ability to predict the (probability of an) outcome of an action makes ARPLACEs a powerful tool when combined with a high level planning system. In this section, we demonstrate how ARPLACE is used in the context of a symbolic transformational planner. Reasoning about ARPLACE enables the planner to generate more robust and efficient plans, and demonstrates the flexibility of the least-commitment ARPLACE representation.

In particular, we consider the task of retrieving two cups from a table. One action sequence that solves this task is: "Plan A: navigate to a location near cup1, pick up cup1 with the left gripper, navigate to a location near cup2, pick up cup2 with the right gripper", as depicted in Figure 16. However, if the cups are sufficiently close to each other (as in Figure 12, right), it is much more efficient to replace the plan above with "Plan B: navigate to a location both near cup1 and cup2, pick up cup1 with the left gripper, pick up cup2 with the right gripper", as it saves an entire navigation action.

---

8. This cost is chosen for its simplicity, to illustrate the generality of the ARPLACE representation. More realistic, complex cost functions can be used.

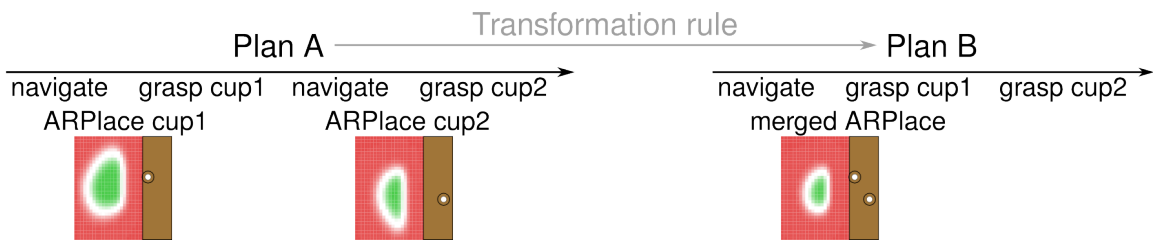

Figure 16: Improving performance through transformational planning with (merged) ARPLACE. Plan A: navigate to two separate poses for grasping each object, using ARPLACEs for both objects. Plan B: navigate to one pose for grasping both objects, using the merged ARPLACE.

Deciding whether to use two base locations (Plan A) or one (Plan B) is difficult to solve in a control program without sacrificing generality. To keep our solution general, we do not want to write two separate control programs for both options, and choose between them with an if-then-else statement. That would mean we have to provide control programs and choice points for every option a robot has. The space of choices is prohibitively large in everyday tasks to allow such an approach. Instead, we use a transformational planner that takes our general program (Plan A) and, if appropriate, applies generic transformation rules that change the program locally (to yield Plan B). Our transformational planner consists of the following components:

**Plan projection.** A projection mechanism for predicting the outcome of a plan. ARPLACE is a compact representation of such a projection mechanism, as it is able to predict the probability of success of an action, given its parameters.

**Flaw detection.** A mechanism for detecting behavior flaws within the predicted plan outcome. Flaws are not only errors that hinder the robot from completing the task, but they may also be performance flaws, such as suboptimal efficiency. Using two navigation actions to approach to cups that are close to each other (Plan A) is flawed, in that it is much more efficient to navigate to one position that is close to both cups.

**Plan transformation.** A mechanism to fix the detected flaws by applying transformation rules to the plan code. For the problem we consider, a local transformation rule is applied to Plan A to yield the more efficient Plan B.

In the next sections, we will describe each of these mechanisms in more detail, and explain how they are implemented to exploit the ARPLACE representation. Note that in this article, we use our transformational planner to exemplify how ARPLACE can be used in the context of a larger planning system. For more information on our transformational planning framework, and further examples of behavior flaws and transformation rules, we refer to the work of Mösenlechner and Beetz (2009).

## 5.1 Plan Design

To detect flaws and apply transformation rules for their repair, the transformational planner must be able to reason about the intention of code parts, infer if a goal has been achieved or not, and deduce what the reason for a possible failure was. To do so, our control programs are written in RPL (McDermott, 1991), which provides functionality for annotating code parts to indicate their purpose and make them transparent to the transformational planner. For the purpose of this article, the most important RPL instructions for semantic annotation in the context of pick-and-place tasks are *achieve*, *perceive* and *at-location*. A formal definition of the semantics of these instructions is given by Mösenlechner and Beetz (2009); here we describe them informally.

*(achieve ?expression)* – If the *achieve* statement executes successfully, the logical expression which is passed as its argument is asserted as being true. For instance, after a successful execution of *(achieve (entity-picked-up ?cup))*, the object referenced by the variable *?cup* must be in the robot's gripper[9].

*(perceive ?object)* – Before manipulating objects, the robot must find the objects and instantiate them in its belief state. After successful execution, the statement *(perceive ?cup)* asserts that the object referenced by *?cup* has been found, and a reference to its internal representation is returned.

*(at-location ?location ?expression)* – Manipulation implies the execution of actions at specific locations. Therefore, it must be assured that pick-up actions are only executed when the robot is at a specific location. *(at-location ?location ...)* asserts that code within its context is either executed at the specified location or fails. Please note that transformations which affect the location where actions are performed directly modify the *?location* parameter of such *at-location* expressions. Therefore, *at-location* is the most important declarative plan expression for optimizing ARPLACEs. To specify locations for *at-location*, we use so-called designators, symbolic descriptions of entities such as locations, objects and actions. For instance, a designator for the location where to stand for picking up a cup can be specified as follows: *(a location (to pick-up) (the object (type cup)))*. This symbolic description is then resolved by reasoning mechanisms such as ARPLACEs and Prolog and an actual pose is generated when it is needed. In general, an infinite number of poses provide a valid solution for such a pose. ARPLACE gives us a way to evaluate their utility and select the best pose.

The declarative expressions explained above can be combined to form a tree. Every *achieve* statement can contain several further *achieve*, *perceive* and *at-location* statements as sub-plans. An example plan tree is sketched in Figure 17. In this tree, the goal *(achieve (entity-at-location ?object ?location))* first perceives the object, then picks it up by achieving *entity-picked-up*, which executes the pick-up action within an *at-location* block, and puts the object down by achieving *entity-put-down*, which also contains an *at-location* block. As we shall see in Section 5.4, behavior flaws are repaired by applying transformation rules that replace sub-trees within the plan tree by new code.

---

9. Please note the lisp syntax, where variables are prefixed with a '?', for example *?cup*, and functions and predicates are pure symbols.

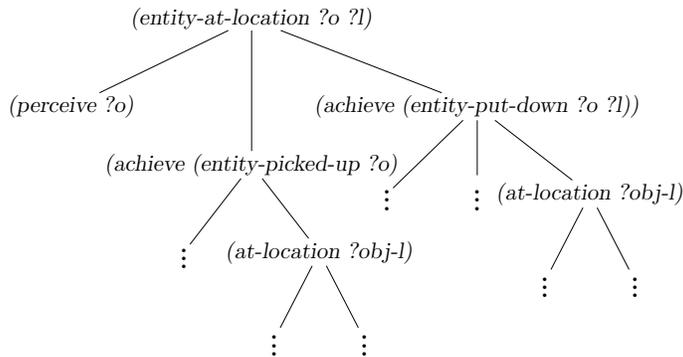

Figure 17: An example plan tree created by executing a pick-up plan

## 5.2 Plan Projection

A central component of a transformational planner is *plan projection*, which simulates the behavior of the robot that arises when executing a plan. In our approach, plan projection generates a temporally ordered set of events based on the plan code presented in the previous section. We use the same Gazebo based mechanism for projection that has been used for generating the training data for learning ARPLACES. In particular, we use information about collisions, perception events and the locations of objects and the robot. While executing the plan in simulation, we generate an extensive execution trace that is then used in our reasoning engine that infers behavior flaws that are then fixed by transformation rules (Mösenlechner & Beetz, 2009). The execution trace contains low-level data representing the position of all objects and the robot, as well as collisions between objects, the visibility of objects for the robot, and information to reconstruct the state of program throughout its execution.

ARPLACES are a very efficient way of performing plan projection, as they predict the probability of a successful outcome *without* requiring on-line generation of execution traces. The reason that execution trace sampling is not required *on-line*, is because the task has already been executed frequently *off-line* during data acquisition (cf. Section 3.1). The results of these task executions have been compiled into the ARPLACES by learning a GSM, which yields a compact representation of the experience acquired. Therefore, this experience must not be generated anew during plan generation.

## 5.3 Behavior Flaws and Reasoning about Plan Execution

Plan projection simulates the robot behavior when executing a plan. The second component of a transformational planner is a reasoning engine that finds pre-defined flaws in the projected robot behavior. Examples of such flaws are collisions, e.g. caused by under-parameterized goal locations, or blocked goals, e.g. when a chair is standing at a location the robot wants to navigate to. The examples above are behavior flaws that lead to critical errors in plan execution (i.e. the plan fails), but we also consider behavior that is inefficient to be flawed (i.e. the plan succeeds, but is unnecessarily inefficient). The task we consider in this paper is an example of such a *performance flaw*, as performing two navigation actions where only one is required is highly inefficient. Behavior flaws are specified using a Prolog-like reasoning engine that is implemented in Common Lisp (Mösenlechner & Beetz, 2009).

The execution trace generated by plan projection is transparently integrated into the reasoning engine, i.e. the execution trace is queried using Prolog predicates. The information recorded in the execution trace is valuable information in order to find behavior flaws. Additional information that is used to find behavior flaws is a set of facts that model the semantics of declarative expressions such as *achieve* or *at-location* and concepts of the world, for instance that objects are placed on "supporting planes" (table, cup-board, ...). To find behavior flaws, their Prolog specifications are matched against the logical representation of the execution trace and if solutions are found, the corresponding flaw is present in the plan and can be fixed.

For instance, the code to match two locations to perform actions that can be merged to one ARPLACE looks as follows:

Listing 1: Flaw definition to match two different pick-up tasks.

```
1  (and
2    (task-goal ?task-1 (achieve (entity-picked-up ?object-1)))
3    (task-goal ?task-2 (achieve (entity-picked-up ?object-2))) (thnot
4    (== ?task-1 ?task-2)) (optimized-action-location ?object-1
5    ?object-2 ?optimized-location))
```

The code above first matches two different pick-up tasks. The predicate *optimized-action-location* holds for *?optimized-location* being an ARPLACE from which the two objects can be picked up. To bind this variable, the predicate is implemented to calculate such an ARPLACE.

Another example for such a flaw definition is failed navigation, i.e. if the robot is not standing at the location it was supposed to drive to:

Listing 2: Flaw definition to find locations that were not reached by the robot although it was told to reach them.

```
1  (and
2    (task-goal ?task (achieve (loc Robot ?goal-loc)))
3    (task-status ?task Done ?t)
4    (holds (loc Robot ?robot-loc) (at ?t))
5    (not (== ?goal-loc ?robot-loc)))
```

The code above first matches the code that is navigating the robot to the location *?goal-loc*. Then it infers the actual location of the robot when the navigation task terminated and binds it to the variable *?robot-loc* and finally asserts that the two locations are not equal. If this Prolog expression can be proven against an execution trace, we have found a flaw indicating an unachieved goal location.

### 5.4 Plan Transformations and Transformation Rules

After a behavior flaw has been detected, the last step of a planner iteration is the application of a transformation rule to fix the behavior flaw. Transformation rules are applied to parts of the plan tree and cause substantial changes in its structure and the corresponding robot behavior.

A transformation rule consists of three parts. The input schema is matched against the plan part that has to be transformed and binds all required code parts to variables in order to reassemble them in the output part. The transformation part performs transformations on the matched parts, and the output plan describes how the new code of the respective plan part has to be reassembled.

$$\frac{\text{input schema}}{\text{output plan}} \Big[ \text{transformation}$$

Besides the integration of ARPLACE into the robot control program through *at-location* statements, ARPLACE is also integrated into the reasoning engine of our transformational planner. Using two locations for grasping is considered a performance flaw if one location would suffice. Informally, we investigate the execution trace for the occurrence of two different pick-up actions, where one is executed at location $L_1$, and the other one is executed at location $L_2$. Then we request a location $L_3$ to perform both actions and the corresponding success probability. $L_3$ is computed by merging the ARPLACE as in Figure 12. If the probability of success of the merged ARPLACE is sufficiently high, we apply a plan transformation, and replace locations $L_1$ and $L_2$ with location $L_3$.

The transformation rule for optimizing ARPLACEs is shown in Listing 3. Please note that all variables that have been bound while matching the flaw definition are still bound and can be used in the transformation rule.

Listing 3: Transformation rule for fixing the flaw.

```
1    (def−tr−rule fix−unoptimized−locations
2      :input−schema
3        ((and (task−goal ?location−task−1
4               (at−location (?location−1) . ?code−1))
5              (sub−task ?location−task−1 ?task−1))
6         (and (task−goal ?location−task−2
7               (at−location (?location−2) . ?code))
8              (sub−task ?location−task−2 ?task−2)))
9      :output−plan
10       ((at−location (?optimized−location) . ?code−1)
11        (at−location (?optimized−location) . ?code−2)))
```

The input schema of the code above consists of two similar patterns, each matching the *at-location* sub-plan of the pick-up goals matched in the flaw. The planner replaces the matching code parts by the corresponding entries of the output plan. In our transformation rule, the location that has been passed to *at-location* is replaced by the optimized location that has been calculated in the flaw definition.

Our behavior flaw is defined to match two different pick-up executions. Then an ARPLACE query is performed to find out the probability for successfully grasping both objects from one location. If the probability is sufficiently high ($> 0.85$) the Prolog query succeeds, i.e. the flaw is detected only if a sufficiently good location for grasping both objects can be found. Note that "sufficiently high" depends very much on the scenario context. In robotic soccer it can be beneficial to choose fast and risky moves, whereas in safe human-robot interaction, certainty of successful execution is more important than mere speed. This article focusses on principled ways of integrating such thresholds in a transformational planner, and relating them to grounded models of the robot's behavior.

What these thresholds should be, and how they are determined, depends on the application domain and the users.

## 6. Empirical Evaluation

In this section we 1) determine how many samples are needed to learn an accurate SVM classifier; 2) compare the robustness of our default strategy for determining base positions with a strategy that uses ARPLACES; 3) compare the efficiency of plans with and without fixing performance flaws with our transformational planner; 4) present preliminary results on the physical robot platform.

### 6.1 Classification Accuracy and Training Set Size

Figure 18 depicts the accuracy of the SVM classifier for predicting which base positions will lead to successful grasps for one particular cup position, evaluated on a separate test set with 150 samples. Without using the capability map to filter out kinematically impossible base positions, the graph levels off after about 300 examples[10]. By filtering out theoretically impossible base positions with the capability map, the classifier achieves the same accuracy within 173 examples (Stulp et al., 2009).

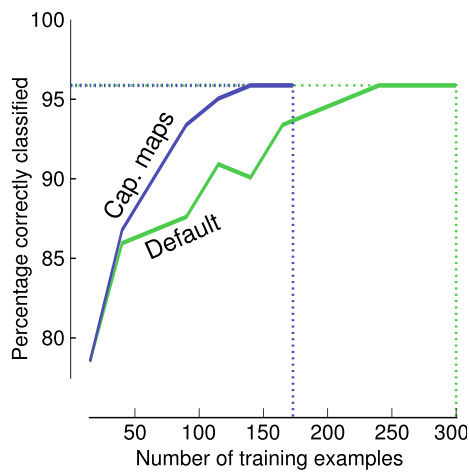

Figure 18: Accuracy dependent on training set size for one cup position.

The effect is more dramatic for the entire dataset containing the data for 16 different cup positions. By applying the capability map, the number of trials that need to be executed reduces from 2992 (all markers in Figure 6) to 666 (only red/green filled markers in Figure 6). As the capability map only reduces unsuccessful attempts, it has no influence on the final classification accuracy, which is 94%.

---

10. This graph applies to another dataset described by Stulp, Fedrizzi, Zacharias, Tenorth, Bandouch, and Beetz (2009), which is very similar to the one used in the rest of this article.

## 6.2 Results from the Simulated Robot

We now compare the robustness of navigation based on probabilistic ARPLACEs with a strategy based on deterministic navigation goals. In this evaluation, the position to which the robot navigates is the position for which ARPLACE returns the highest probability that grasping the target object will succeed. We compare this strategy to our previous hand-coded implementation FIXED, which always navigates to a location that has the same relative offset to the target object, whilst at the same time taking care not to bump into the table.

In these experiments, we vary the position of the cup $(\Delta x^{obj}, \Delta \psi^{obj})$, as well as the uncertainties the robot has about its own position and the position of the cup, by varying the diagonal elements of the covariance matrices associated with the position of the robot $(\sigma_{\Delta x^{rob} \Delta x^{rob}}, \sigma_{\Delta y^{rob} \Delta y^{rob}})$ and the cup $(\sigma_{\Delta x^{obj} \Delta x^{obj}}, \sigma_{\Delta \psi^{obj} \Delta \psi^{obj}})$. For each combination of these variables, the robot performs the navigate-reach-grasp-lift sequence. The result is recorded, just as during data acquisition for learning the Generalized Success Model. To simulate the uncertainty, we sample a specific perceived robot and cup position from the distribution defined by their means and covariance matrices. The result of the action is determined by the true simulated state of the world, but the robot bases its decisions on the perceived samples.

The results of this evaluation are summarized in the three bar plots in Figure 19, which depict the success ratios of the ARPLACE-based and FIXED strategies. Each ratio is the number of successful executions, divided by the number of examples, which is 100. The $p$-value above each pair of bars is computed with a $\chi^2$ test between them, which tests whether the number of successful and failed attempts is sampled from the same distribution for ARPLACE and FIXED.

The first graph depicts the success ratios for increasing uncertainty about the object position (i.e. $\sigma_{\Delta x^{obj} \Delta x^{obj}} = [\ 0.00\ 0.05\ 0.10\ 0.15\ 0.20\ ]$), for fixed robot position uncertainty $\sigma_{\Delta x^{rob}} = 0.05$. In all cases, the ARPLACE strategy significantly outperforms the FIXED strategy. Furthermore, the performance of ARPLACE is much more robust towards increasing object position uncertainty, as ARPLACE takes this explicitly into account.

The same trend can be seen when increasing the uncertainty in the robot position (i.e. $\sigma_{\Delta x^{rob} \Delta x^{rob}} = \sigma_{\Delta y^{rob} \Delta y^{rob}} = [\ 0.00\ 0.05\ 0.10\ 0.15\ 0.20\ ]$), for fixed object position uncertainty $\sigma_{\Delta x^{obj}} = 0.05$. However, when $\sigma_{\Delta x^{rob} \Delta x^{rob}} > 0.1$ the difference between ARPLACE and FIXED is no longer significant.

Finally, the last graph depicts the success ratios when increasing both robot and object uncertainty. Again, ARPLACE significantly outperforms FIXED when ($\sigma < 0.15$). If the robot is quite uncertain about its own and the object's position ($\sigma > 0.15$), grasp success probabilities drop below 50% for both strategies.

Summarizing, ARPLACE is more robust towards state-estimation uncertainties than our previous default strategy. The effect is more pronounced for object positions than robot positions.

## 6.3 Transformational Planning with Merged ARPlaces

We evaluated the merging of ARPLACEs for joint grasping, and the application of transformation rules with our RPL planner, as discussed in Section 4.4.1. Two cups are placed on the table, where the distance between them is varied between 20 and 60cm, with increments of 5cm. Our evaluation shows that grasping two cups from separate base positions requires on average 48 seconds, independent of the relative distance of the cups to each other. By applying transformation rules, the default plan is optimized to 32 seconds, which is a significant ($t$-test: $p < 0.001$) and substantial performance gain of 50% (Fedrizzi et al., 2009). Above 45cm, two cups cannot be grasped from one position, and plan transformation is not applied.

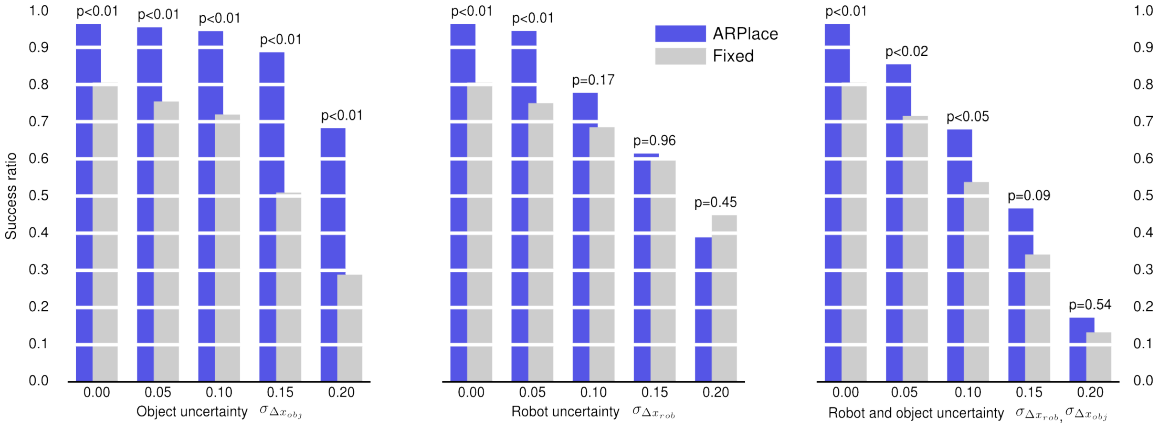

Figure 19: Success ratios of the ARPLACE and FIXED approaches when changing object and/or robot pose uncertainties.

### 6.4 Integration of ARPLACE in the Physical Robot System

At a day of open house, our B21 mobile manipulation platform continually performed an application scenario, where it locates, grasps, and lifts a cup from the table and moves it to the kitchen oven. Figure 20 shows two images taken during the demonstration. The robot performed this scenario 50 times in approximately 6 hours, which has convinced us that the robot hardware and software are robust enough to be deployed amongst the general public.

After the open day, we ran the same experiment, but this time we determined the goal location for navigating to the table as being the mode of the ARPLACE that was computed before executing the navigation action. Since the main focus of this experiment was on our error-recovery system described by Beetz et al. (2010), the improved robot performance we observed cannot quantitatively be attributed to the use of ARPLACE or the error-recovery system. However, a major qualitative improvement we certainly can attribute to using ARPLACE was that the cup can now be grasped from a much larger area on the table. Without ARPLACEs, the cup always had to be placed on the same position on the table to enable successful grasping.

## 7. Conclusion

In this article, we present a system that enables robots to learn action-related places from observed experience, and reason with these places to generate robust, flexible, least-commitment plans for mobile manipulation. ARPLACE is modeled as a probability distribution that maps locations to the predicted outcome of an action.

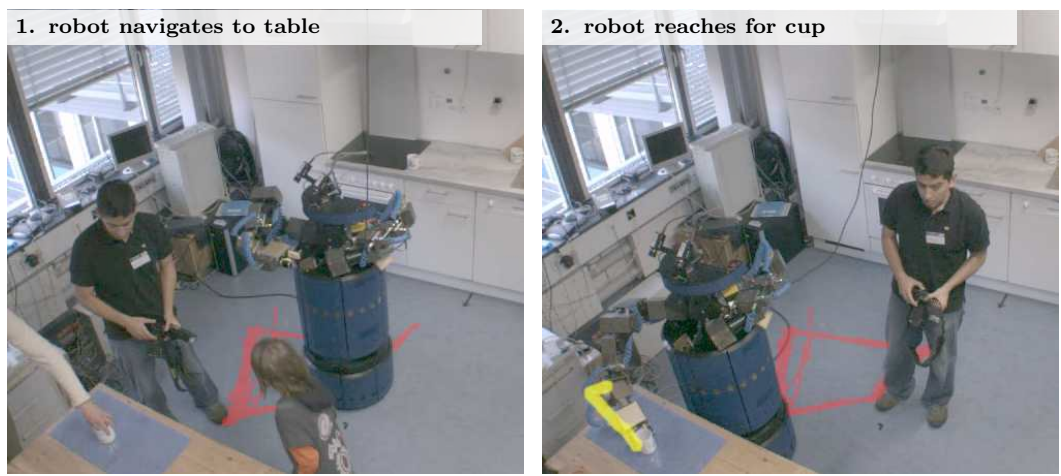

Figure 20: A reach and grasp trajectory performed during a public demonstration. (Note that the operator is holding a camera, not a remote control!)

We believe our system has several advantages. First of all, the learned model is very compact, with only 2 (deformation) parameters, which are directly related to task-relevant parameters. Querying the model on-line is therefore very efficient. This is an advantage of compiling experience into compact models, rather than running a novel search for each situation.

On the other hand, as the model is acquired through experience-based learning, the model is grounded in observed experience, and takes into account the robot hardware, its control programs, and interactions with the environment. It can be applied to any mobile manipulation platform, independent of the manipulators, navigation base, or the algorithms that run on them.

The output of this model is a set of positions with associated success probabilities, instead of one specific position. Rather than constraining itself to a specific position prematurely, the robot can efficiently update ARPLACE as new sensor data comes in. This enables least-commitment planning. The ARPLACE representation also enables the optimization of secondary criteria, such as execution duration, or determining the best position for grasping two objects simultaneously. In previous work, we proposed *subgoal refinement* (Stulp & Beetz, 2008) for optimizing such secondary criteria with respect to subgoals.

Finally, by using ARPLACEs to determine appropriate base positions, difficult positions for grasping are avoided, which leads to more robust behavior in the face of state estimation uncertainty, as demonstrated in our empirical evaluation.

We are currently extending our approach in several directions. We are in the process of including ARPLACE in a more general utility-based framework, in which the probability of success is only one of the aspects of the task that needs to be optimized. New utilities, such as execution duration or power consumption, are easily included in this framework, which enables the robot to trade off efficiency and robustness on-line during task execution.

We are also applying our approach to more complex scenarios and different domains. For instance, we are learning higher-dimensional ARPLACE concepts, which take more aspects of the scenario into account, i.e. different object sizes and objects that require different types of grasps. Instead of mapping specific objects to places, we will map object and grasp *properties* to deformation modes. We are also investigating extensions and other machine learning algorithms that will enable our methods to generalize over this larger space. Objects which require very different grasps, such as using two hands to manipulate them, will require more sophisticated methods for acquiring and reasoning about place. Generalization of our place concept with respect to situations and task contexts is a research challenge which we have on our mid-term research agenda.

## Acknowledgments

We are grateful to Pierre Roduit for providing us with the Matlab code described by Roduit et al. (2007). We also thank Ingo Kresse, Alexis Maldonado, and Federico Ruiz for assistance with the robotic platform, and the robot system overview. We are grateful to Franziska Zacharias for providing a capability map (Zacharias et al., 2007) for our robot. We thank Dominik Jain and Franziska Meier for fruitful discussions on Section 4.2.

This work was partly funded by the DFG project ActAR (Action Awareness in Autonomous Robots) and the CoTeSys cluster of excellence (Cognition for Technical Systems, http://www.cotesys.org), part of the Excellence Initiative of the German Research Foundation (DFG). Freek Stulp was also supported by a post-doctoral Research Fellowship (STU-514/1-1) from the DFG, as well as by the Japanese Society for the Promotion of Science (PE08571). Freek Stulp's contributions to this work were made at the Intelligent Autonomous Systems Group (Technische Universität München, Munich, Germany), the Computational Neuroscience Laboratories (Advanced Telecommunications Research Institute International, Kyoto, Japan), and the Computational Learning and Motor Control Lab (University of Southern California, Los Angeles, USA),

## Appendix A. Robot Platform

The action sequence we consider in this article is: 1) navigate to a specified base position near the table; 2) reach for the object; 3) close the gripper; 4) lift the object. We now sequentially describe the various hard- and software components involved in executing these actions. An overview of these components and the data communicated between them is depicted in Figure 21.

The main hardware component is a B21r mobile robot from Real World Interfaces (RWI), with a frontal 180 degrees Sick LMS 200 laser range scanner. Before task execution, the robot acquires a map of the (kitchen) environment using 'pmap' for map building. To navigate to the specified base position, the robot uses an Adaptive Monte Carlo Localization algorithm for localization, and the AMCL Wavefront Planner for global path planning. For these three software modules (map building, localization and planning), we use the implementations from the Player project (Gerkey et al., 2003).

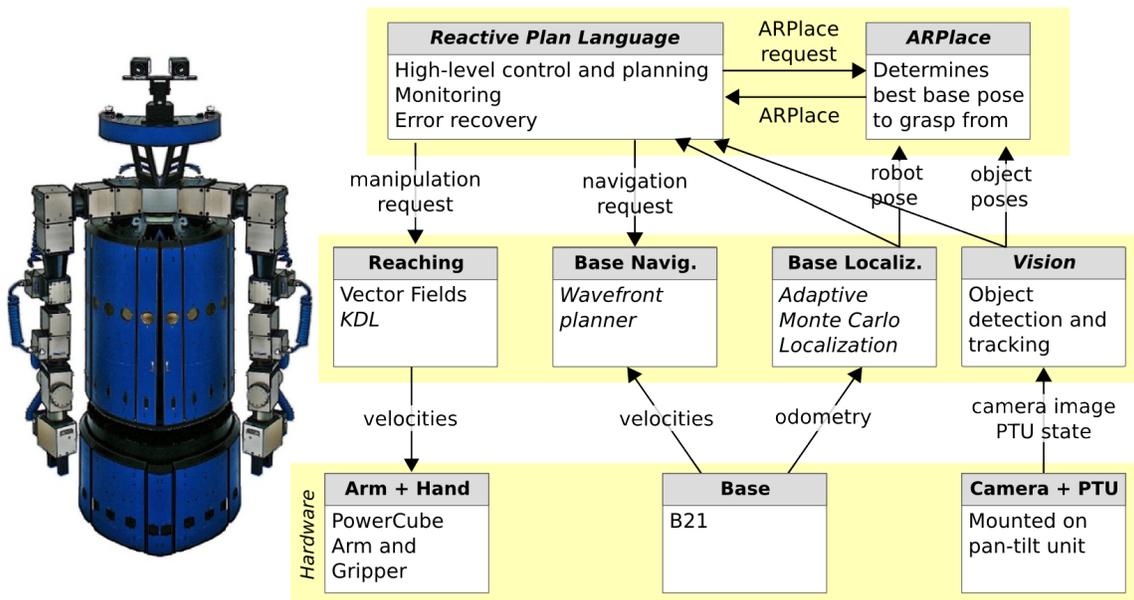

Figure 21: Overview of mobile manipulation hardware and software modules.

When the robot is close to the table, it detects and tracks the target object using the approach proposed by Klank et al. (2009). The stereo-vision hardware consists of two high dynamic range cameras that are mounted on a PTU-46 pan-tilt unit from Directed Perception and have a resolution of 1390x1038 pixels.

For manipulation, the robot is equipped with two 6-DOF Powercube lightweight arms from Amtec Robotics. To control the arms and reach for the target cup, we use the Kinematics and Dynamics Library (Orocos-KDL) (Smits, ) and a Vector Field approach. Within this vector field, the handle of the cup is an attractor, but the cup itself, the table and all other obstacles are repellors. Details about the position and shape of these attractors and repellors are given by Beetz et al. (2010). On-line at every control cycle, the task space velocity at the end-effector is computed given the attractors and repellors, and this velocity is mapped to joint space velocities using a damped least squares inverse kinematics algorithm. After reaching the desired end-effector pose, the 1-DOF slide gripper closes.

High-level decision making, monitoring and error-recovery is done by the planning module written in the Reactive Planning Language (McDermott, 1991). It requests ARPLACEs from the module described in this article, reasons about them, and performs navigation and manipulation requests based on them.

Communication between all modules described above is done over a middleware layer consisting of Player (Gerkey et al., 2003) and YARP (Metta, Fitzpatrick, & Natale, 2006). This overview is a simplification of the actual system. For instance, the role of RFID tags and the Belief State have been omitted. For a complete and more detailed description of the mobile manipulation platform, we refer to the work of Beetz et al. (2010, Section 1.2).

## Appendix B. Landmark Distribution for the Point Distribution Model

A Point Distribution Model (PDM) takes a set of $m$ landmarks on $n$ contours as an input, represented as a $m \times n$ matrix $\mathbf{H}$, and returns the matrices $\overline{\mathbf{H}}$ (mean of the contours), $\mathbf{P}$ (deformation modes), and $\mathbf{B}$ (deformation mode weighting per contour), which the original contours can be reconstructed.

In our application of PDMs, we are free to choose the locations of the landmarks. Therefore, the goal of the procedure described here is to determine landmark locations that leads to a compact PDM that accurately reconstructs the original contours, i.e. the classification boundaries. We do so by explicitly optimizing two measures: 1) model compactness: the amount of energy $e$ stored in the first $d$ degrees of freedom of the PDM, with $0 \leq e \leq 1$; 2) reconstruction accuracy: the mean distance $l$ between the landmarks on the original contours and reconstructed contours. These measures are combined in the cost function $(2 - e)l^2$, expressing that we want low error and high energy for a given number of degrees of freedom $d$.

Given the number of landmarks $m$ and the number of degrees of freedom $d$, we explicitly optimize this cost function through search. We do so by varying the position of each landmark, one landmark at a time, and greedily selecting the position that leads to the lowest cost. This optimization is first done for $d = 1$, and the number of degrees of freedom $d$ is incremented until the optimization leads to an energy that lies above 95%. This ensures that the number of degrees of freedom $d$ and the distance $l$ between the landmarks remains low, whilst the energy $e$ is high. Therefore, the resulting PDM model will be compact yet accurate.

This optimization step is by far the most computationally intensive step in the off-line learning phase. We are currently investigating the use of alignment methods from computer vision (Huang, Paragios, & Metaxas, 2006), to replace our iterative optimization approach.

## Appendix C. From Robot Coordinate Systems to the Relative Feature Space

Our robot uses a variety of coordinate systems. The goal is to compute the matrix ${}^{\text{GSM}}T_O$, which describes the object's position relative to the feature space of the Generalized Success Model. ${}^{\text{GSM}}T_O$ can then be used to reconstruct classification boundaries for successfully grasping the object, as described in Section 3.4. We now present the required coordinate systems, and how they are transformed to yield the Generalized Success Model required feature space that is depicted in Figure 5.

The coordinate frames that are involved in the transformation are depicted in Figure 22: the world frame $F_W$, the table frame $F_T$ that is centered in the middle top of the table, the robot frame $F_R$ that is centered in the robot's base center at the floor, the camera frame $F_C$ that is centered in the camera's sensor chip, the frame of the pan-tilt unit where the camera is mounted $F_{PT}$, and the relative feature space $F_{\text{GSM}}$.

To acquire the position of the target object $O$ relative to $F_{\text{GSM}}$, we compute ${}^{\text{GSM}}T_O$ as follows:

$$ {}^{\text{GSM}}T_O \;=\; ({}^W T_{\text{GSM}})^{-1} \; * \; {}^W T_O \tag{12} $$

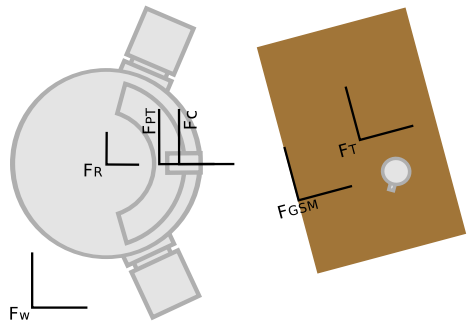

Figure 22: Relevant coordinate frames

The global position of the object $^{W}T_{O}$ is computed as follows:

$$^{W}T_{O} \; = \; ^{W}T_{R} \; * \; ^{R}T_{PT} \; * \; ^{PT}T_{C} \; * \; ^{C}T_{O} \tag{13}$$

Here, $^{W}T_{R}$ is the location of the robot's base frame relative to the world frame. The robot uses an AMCL particle filter for estimating its position. $^{R}T_{PT}$ is the pose of the pan tilt unit relative to the robot's base frame. The transformation matrix $^{R}T_{PT}$ is constant and was specified by manually measuring the distances and angular offsets from the B21 robot base to the pan tilt unit. Because of careful measurement, we assume maximum errors of 1mm for the distance measurements along the x-, y-, and z-axis, and 2° for the yaw angle measurement. $^{PT}T_{C}$ is the pose of the camera's sensor relative to the pan tilt unit. $^{PT}T_{C}$ changes according to the current pan and tilt angles, but can be read from the pan tilt unit's driver with high accuracy. $^{C}T_{O}$ is the position of the target object relative to the camera frame. It is estimated by the vision-based object localization module that is described by Klank et al. (2009).

In order to compute $^{W}T_{\text{GSM}}$ we need to know the global position of the object $^{W}T_{O}$, which we already computed above, and the global position of the table $^{W}T_{T}$. Currently, we get the world coordinates of the table's position from a map, but it is also possible to estimate its position with the vision-based object localization module. We then compute the normal from the object to the table edge that is closest to the robot, as can be seen in Figure 5. The origin of $F_{\text{GSM}}$ and therefore $^{W}T_{\text{GSM}}$ is where the table edge and the object normal intersect.

The most critical parts in the computations above are the angular estimations of the robot's localization and vision system. First, their estimation uncertainty is rather big. Second, an error in the localization angle has a significant impact on the estimated object pose, as follows from Equation 13.

The pose of the cup in frame $F_{\text{GSM}}$ is a 6D vector $[x, y, z, yaw, pitch, roll]$. However, since we assume the cup is standing upright on the table, we set $z$ to the table's height, and $roll$ and $pitch$ to 0°. Since the origin of $F_{\text{GSM}}$ is perpendicular to the table's edge that passes through $y$, $y$ is also 0 by definition. The remaining parameters $x$ and $yaw$ then correspond to the features $\Delta x^{obj}$ and $\Delta \psi^{obj}$ respectively.